\documentclass[journal]{IEEEtran}
\usepackage{color}
\usepackage{multirow}
\usepackage{bigstrut}
\usepackage{graphicx}
\usepackage{subfigure}
\usepackage{amssymb}
\usepackage{amsmath}
\usepackage{url}
\ifCLASSINFOpdf
\else
\fi
\usepackage{graphicx}
\usepackage{gensymb}
\begin{document}
%
\title{Distortion-Aware Loop Filtering of Intra 360$^o$ Video Coding with Equirectangular Projection}

%
\author{Pingping~Zhang,
        Xu~Wang,~\IEEEmembership{Member,~IEEE,}
        Linwei~Zhu,~\IEEEmembership{Member,~IEEE,}
        Yun~Zhang,~\IEEEmembership{Senior Member,~IEEE,}
        Shiqi~Wang,~\IEEEmembership{Member,~IEEE,}
        Sam Kwong,~\IEEEmembership{Fellow,~IEEE}
\thanks{This work was supported in part by the National Natural Science Foundation of China (Grant 61871270 and Grant 61672443), in part by the Supported by Shenzhen Fundamental Research Program (grant no. JCYJ20200109110410133 and JCYJ20200812110350001) and in part by the National Engineering Laboratory for Big Data System Computing Technology of China.

Pingping Zhang and Xu Wang are with the College of Computer Science and Software Engineering, Shenzhen University, China, and also with Guangdong Laboratory of Artificial Intelligence and Digital Economy (SZ), Shenzhen University, Shenzhen, 518060, China. Email: (ppingyes@gmail.com, wangxu@szu.edu.cn).

Linwei Zhu and Yun Zhang are with the Shenzhen Institutes of Advanced Technology, Chinese Academy of Sciences, Shenzhen 518055, China. Email: (lwzhu2-c@my.cityu.edu.hk, yun.zhang@siat.ac.cn).

Shiqi Wang and Sam Kwong are with the Department of Computer Science, City University of Hong Kong, Kowloon, Hong Kong. Email: (shiqwang@cityu.edu.hk; cssamk@cityu.edu.hk).
}
}
\markboth{SUBMITTED TO IEEE TRANSACTIONS ON CIRCUITS AND SYSTEMS FOR VIDEO TECHNOLOGY, June. 2021}%
{Shell \MakeLowercase{\textit{et al.}}: Bare Demo of IEEEtran.cls for IEEE Journals}
%



\maketitle


\begin{abstract}
In this paper, we propose a distortion-aware loop filtering model to improve the performance of intra coding for 360$^o$ videos projected via equirectangular projection (ERP) format. To enable the awareness of distortion, our proposed module analyzes content characteristics based on a coding unit (CU) partition mask and processes them through partial convolution to activate the specified area. The feature recalibration module, which leverages cascaded residual channel-wise attention blocks (RCABs) to adjust the inter-channel and intra-channel features automatically, is capable of adapting with different quality levels. The perceptual geometry optimization combining with weighted mean squared error (WMSE) and the perceptual loss guarantees both the local field of view (FoV) and global image reconstruction with high quality. Extensive experimental results show that our proposed scheme achieves significant bitrate savings compared with the anchor (HM + 360Lib), leading to 8.9\%, 9.0\%, 7.1\% and 7.4\%  on average bit rate reductions in terms of PSNR, WPSNR, and PSNR of two viewports for luminance component of 360$^o$ videos, respectively.
\end{abstract}


%
\IEEEpeerreviewmaketitle
\section{Introduction}
\IEEEPARstart{R}{ecently}, immersive media with three degrees of freedom (3DoF), \textit{i.e.}, 360$^o$ video, has reached sufficient maturity for both industrial application and standardization. Users can look around at a fixed viewing position for the captured scene with the aid of real-time interactive devices such as a head-mounted display (HMD). To provide users with an amazing immersive experience, 360$^o$ videos need much more bandwidth compared with conventional 2D videos due to the requirements of both high resolution and frame rate~\cite{lin19jestcs}. For example, the resolution of the popular virtual reality (VR) headset Oculus Rift is $1080 \times 1200$ with 110$^o$ field of view (FoV), which requires the captured panoramic video with the resolution of 4K and beyond. Therefore, efficient compression methods are highly desired for storing and transmitting such a massive amount of data.

The common 360$^o$ scene acquisition approach is to record multi-view videos around the observer through multiple cameras simultaneously. After a series of procedures, \textit{e.g.,} image stitching, projection and mapping, visual information from an original 360$^o$ scene is mapped from a 3D spherical surface into a 2D plane, which can be directly fed into the conventional video system for compression without modification of the core codec~\cite{Wien19jestcs}. However, the projection and mapping will cause deformation (or distortion) of video contents. Existing works mainly focus on developing projection formats and advanced tools to improve the coding performance. Specifically, the Joint Collaborative Team on Video Coding (JCT-VC) and Joint Video Experts Team (JVET) are working on the standardization of 360$^o$ video compression as the extension of High Efficiency Video Coding (HEVC)~\cite{sullivan2012overview} and Versatile Video Coding (VVC) \cite{yang2019low}, respectively.

Due to the design philosophy of the block-based video coding framework adopted in HEVC, the quantization stage after prediction and transformation inevitably causes serious distortions, \textit{e.g.,} blurring, blocking and ringing artifacts. Consequently, the reconstructed video quality is significantly degraded, especially in the scenario with low bandwidth constraints. To improve the visual quality of reconstructed videos, numerous efforts have been made by investigating in-loop filtering techniques. For example, to reduce blocking artifacts, the deblocking filter (DF)~\cite{list2003adaptive}~\cite{ norkin2012hevc} first analyzes the distortion degree of each coding unit (CU), and then performs filtering adaptively. The sample adaptive offset (SAO)~\cite{fu2012sample} reduces sample distortion by classifying reconstructed samples into different categories. For each category, an offset is obtained at the encoder, encoded and transmitted in the bitstream, and compensated at the decoder. To further remove the artifacts caused by block-based processing and coarse quantization, other types of in-loop filtering techniques such as adaptive loop filtering (ALF)~\cite{tsai2013adaptive} and nonlocal in-loop filtering~\cite{Ma16MM,zxf17tcsvt} have also been proposed.

Inspired by the success of convolutional neural network (CNN) in solving image restoration tasks, CNN-based in-loop filtering techniques~\cite{zhang2018residual, jia2019content,zhang19tcsvt} have been widely investigated. However, most existing approaches are mainly designed for the perspective image on the regular 2D grid only~\cite{maNNtcsvt}. Since the projection stage of 360$^o$ video system is non-uniform, \textit{e.g.}, the equirectangular projection (ERP) over-sampled the poles than the equator, kernel sizes or shapes of convolutional layers should be adapted to the latitude~\cite{pmlr19khasanova19}. Due to the limitation of a plain CNN model with fixed kernel size, directly applying existing in-loop filtering models on 360$^o$ video coding can not achieve promising performance. Recently, several attempts have been conducted for tackling 360$^o$ image/video analysis tasks~\cite{xu2020state} such as geometry-aware CNN models ~\cite{pmlr19khasanova19,tateno2018distortion,su2019kernel}. Since the data locality of operation layers may be broken by the operations of geometry-aware CNN models, \textit{e.g.}, grid sampling, the inference speed and memory requirements will become the bottleneck of the 360$^o$ image restoration tasks.
 
In this paper, we propose a loop filtering model for intra 360$^o$ video coding via a distortion-aware CNN architecture, aiming to improve the performance of 360$^o$ video coding by enhancing the reconstruction quality of intra coded equirectangular images. First, the proposed model extracts representative feature maps from the input distorted frame, which analyzes the content to conduct the partial convolution with different kernel sizes. Subsequently, the feature recalibration module is utilized to re-weight each feature channel to adapt different quality features. Finally, the image reconstruction module reconstructs and outputs the restored frame. Experimental results demonstrate that our proposed model can well depict the geometry characteristics of the equirectangular images, yielding a significant improvement in coding performance. Overall, the main contributions of this paper are summarized as follows:
\begin{itemize}
    \item The proposed CU-based feature extraction module can analyze contents of the equirectangular image, and approximately activate different contents via partial convolutions with different kernel sizes.
    \item To adapt videos with various compression qualities, we propose a feature recalibration module via cascaded residual channel-wise attention blocks (RCABs), which can automatically re-adjust the weight of inter-channel and intra-channel features.
    \item To guarantee both local FoV and global image qualities of the restored equirectangular image, the proposed distortion-aware optimization combines the weighted mean squared error (WMSE) and the perceptual loss for the model training.
\end{itemize}

The remainder of this paper is organized as follows. Sec.~\ref{sec:background} overviews the related works on CNN-based filtering and deep learning for 360$^o$ image/video processing. In Sec.~\ref{sec:alg}, the proposed distortion-aware loop filtering model for intra 360$^o$ video coding is described in detail. Sec.~\ref{sec:exp} introduces the experimental results. Finally, the conclusion is drawn in Sec.~\ref{sec:con}.

\begin{figure*}[htbp]
    \centering
    \includegraphics[width=1.0\textwidth]{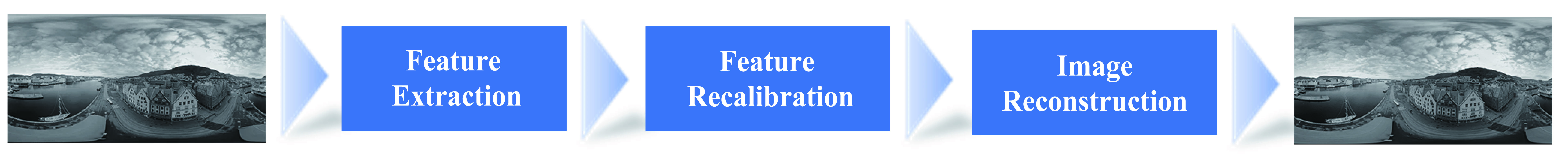}
    \caption{The proposed model of enhancing the quality of reconstructed intra frame. The kernel size adaption with respect to both the evaluation and image content is considered in our proposed feature extraction module. Besides, the feature maps is adaptively re-weighted with respect to QP setting in our proposed feature recalibration module. Afterwards, a loop filtering network, which is trained by considering the distortion-aware loss in terms of WMSE, is used to produce the corresponding restoration results.}
    \label{fig:model}
\end{figure*}

\section{Related Works}
\label{sec:background}
\subsection{CNN-based Filtering}

According to the integration way in the video codec, existing CNN-based filtering models for compression artifacts removal can be classified into two categories~\cite{dlvc}: post-filtering and in-loop filtering.

\textbf{CNN-based post-filtering model} for compression artifacts removal can be treated as an image restoration task, which does not need to modify the codec architecture and directly enhances the reconstructed video quality at the decoder side. For HEVC intra coding, Dai \textit{et al.} proposed the VRCNN model~\cite{dai2017convolutional} which includes four layers and adopts variable filter sizes and residual connections for different layers. Yang \textit{et al.}~\cite{yang2018enhancing} proposed the quality enhancement CNN (QE-CNN) architecture for both HEVC intra- and inter-coding frames, which can maximize the quality enhancement under the constraint of computational time through their proposed TQEO scheme. To utilize the temporal correlation among frames, the multi-frame quality enhancement (MFQE)~\cite{yang2018cvpr} scheme was proposed to enhance the low-quality frames under the guidance of their neighboring frames with peak qualities. However, the performance gains achieved by these models may be limited for low bitrate application scenarios due to the severe structural distortion caused in the quantization stage.

The HEVC bitstream contains implicity content priors such as motion vectors and block coding modes, which can be exploited for guiding the quality enhancement of decoded frames. For example, Ma \textit{et al.}~\cite{ma2018residual} proposed to feed both predicted residual and decoded frame into the CNN. Since a block-based compression mechanism mainly causes the compression artifacts, the decoded image and its associated block partition information are integrated into a CNN-based filtering model to guide the quality enhancement. For example, Kang \textit{et al.}~\cite{kang2017multi} utilized only CU and transform unit (TU) information of each frame, and Lin \textit{et al.}~\cite{lin2019partition} proposed to utilize CU size information for mask generation and mask-patch fusion. Instead of mask-frame fusion~\cite{lin2019partition}, our work utilizes the CU partition information to guide feature extraction to achieve kernel size adaption.

\textbf{CNN-based in-loop filtering model} is more challenging and aims to be integrated into the coding loop. Since the filtered frame with improved quality will be referenced by the following frames, the R-D performance can be directly improved. According to the integration position of CNN based models, existing works can be divided into three classes: 1) Directly replacing the conventional in-loop filter module in the video codec, \textit{e.g.}, the IFCNN \cite{park2016cnn}, and MDCNN~\cite{kuanar2018deep} architectures were proposed to replace the SAO module for P/B frames. 2) Filtering following the DF and SAO modules, \textit{e.g.}, RHCNN~\cite{zhang2018residual} and STResNet~\cite{jia2017spatial}. The RHCNN model consists of several cascaded residual highway units and shortcuts, while the STResNet is a spatial-temporal residual network. 3) Inserting the CNN-based filtering model between the DF and SAO modules, \textit{e.g.}, MLSDRN~\cite{meng2018new} and CNN-ILF~\cite{dlvc}.

Different from the image restoration task that only aims to maximize global image quality, the CNN-based in-loop filtering models also need to address some systematic issues~\cite{ding2019switchable}, such as the local quality of coding tree units (CTUs) that may be referenced by the following frames, the trade-off between the computational complexity of inference and coding efficiency. To achieve superiority in both computational complexity and coding efficiency, Ding \textit{et al.}~\cite{ding2019switchable} integrated the proposed SEFCNN into the video codec by adopting a switchable mechanism between the CNN-based and the conventional in-loop filter modules, which enabled enhancement selectively from frame level to CU level and avoided the double enhancing effect. Jia \textit{et al.}~\cite{jia2019content} proposed a content-aware loop filtering scheme based on multiple CNN models, which adaptively selected the CNN model for each CTU according to the class label output by the content analysis network. The model inference was controlled by signalling frame-level and CTU-level flags. Above mentioned approaches train and store separate models for each QP band, which are inefficient and impractical. To tackle the aforementioned problem, the RRCNN model~\cite{zhang19tcsvt} was designed to train a single model for different bitrate settings with the guidance of QP maps.

Overall, previous CNN-based filtering works are mainly designed for perspective images with regular sampling grids, and there are rare works proposed for considering the characteristics of equirectangular images/videos, one of the most common 360$^o$ data representation formats. In this paper, our work will measure the distorion-aware loss to guide the filtering process, to achieve improvement on both local FoV and global image qualities. Besides, the proposed feature recalibration module can train a single model for various QP settings.

\subsection{Deep Learning for 360$\degree$ Image/Video Processing} Recently, numerous deep learning-based models have been proposed for tackling challenging 360$^o$ image/video analysis tasks~\cite{xu2020state}, \textit{e.g.}, visual attention models, object detection and quality assessment for 360$^o$ contents. Due to the irregular geometric mapping, directly applying plain CNN models on the equirectangular image/video may not work well.

Theoretically, the best solution of CNN-based 360$^o$ image/video processing is to take advantage of the image projective geometry priors for network architecture designing by adaptively adjusting the size or shape (position offset) of convolutional kernels~\cite{pmlr19khasanova19}. To utilize domain adaptation of CNNs from perspective images to equirectangular image, Su \textit{et al.}~\cite{su2017learning} firstly trained separate models with varied convolutional kernel sizes, which are adapted to different latitudes of the equirectangular images at the cost of large model parameters. Zhao \textit{et al.}~\cite{zhao2018distortion} proposed a distortion-aware CNN model for spherical images to account for varying distortion effects. Specifically, their model needs to sample the points on the tangent plane uniformly, then back-projects these points to the sphere to determine the kernel offsets. Similarly, Tateno \textit{et al.}~\cite{tateno2018distortion} proposed distortion-aware convolutions for dense 360$^o$ depth map prediction. Later, Su \textit{et al.}~\cite{su2019kernel} proposed a kernel transformer network (KTN) for transferring CNNs from perspective images to equirectangular images by learning a kernel transform function for the projective geometry.

However, it is hard to apply these geometry-aware CNN models on image restoration tasks directly. The reasons are two-fold. First, the grid sampling of distortion-aware CNN models will breakdown the data locality of operation layers and significantly slow the forward inference speed as the resolution increases. Second, the geometry-aware CNN-based models need to process the whole frame together, and the GPU memory consumption will be unaffordable for an equirectangular image with a resolution larger than 4K. To avoid massive consumption of computational resources and improve the adaption of kernels, we utilize the CU partition information to process different contents in the proposed feature extraction module.

Meanwhile, recent years have witnessed the growing interest in researches on viewport-depended processing for 360 content~\cite{xu2020state} because viewers focus on a small part of the whole 360$\degree$ image and quality degradation of viewport images is much noticeable. Many research areas, \textit{e.g.,} saliency prediction~\cite{qiao2020viewport} and visual quality assessment~\cite{li2019viewport} have carried on researches on the viewport images. Motivated by these works, we introduce the perceptual loss to improve the quality of viewport images.

\section{Proposed Distortion-aware Loop Filtering}
\label{sec:alg}

According to the characteristic of $360\degree$ videos, our proposed distortion-aware model consists of a feature extraction module, a feature recalibration module and an image reconstruction module, shown in Fig.~\ref{fig:model}. Our proposed module first analyzes content characteristics, activates and processes specified areas. The feature recalibration module then leverages cascaded RCABs to adjust the inter-channel and intra-channel features automatically. Thus, it is capable of adapting with different qualities. Finally, the feature maps are mapped into a restored frame via an image reconstruction module. Details on the modules designing, distortion-aware loss function, and implementation of the loop filtering are provided as follows.

\begin{figure}[t]
\subfigure[]{
\centering
\includegraphics[width=0.98\linewidth]{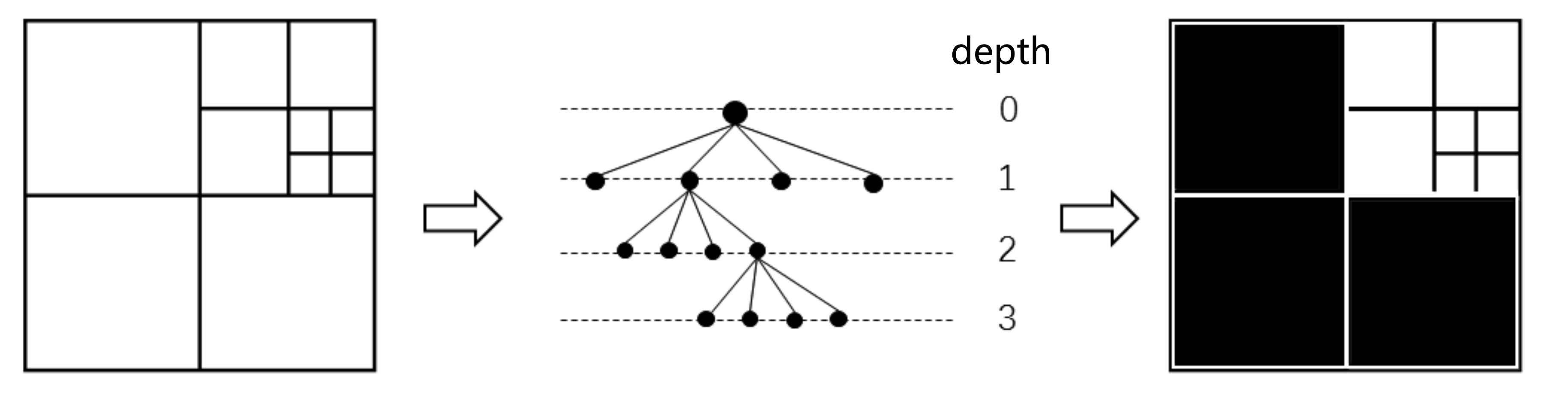}
}
\subfigure[]{
\centering
\includegraphics[width=0.98\linewidth]{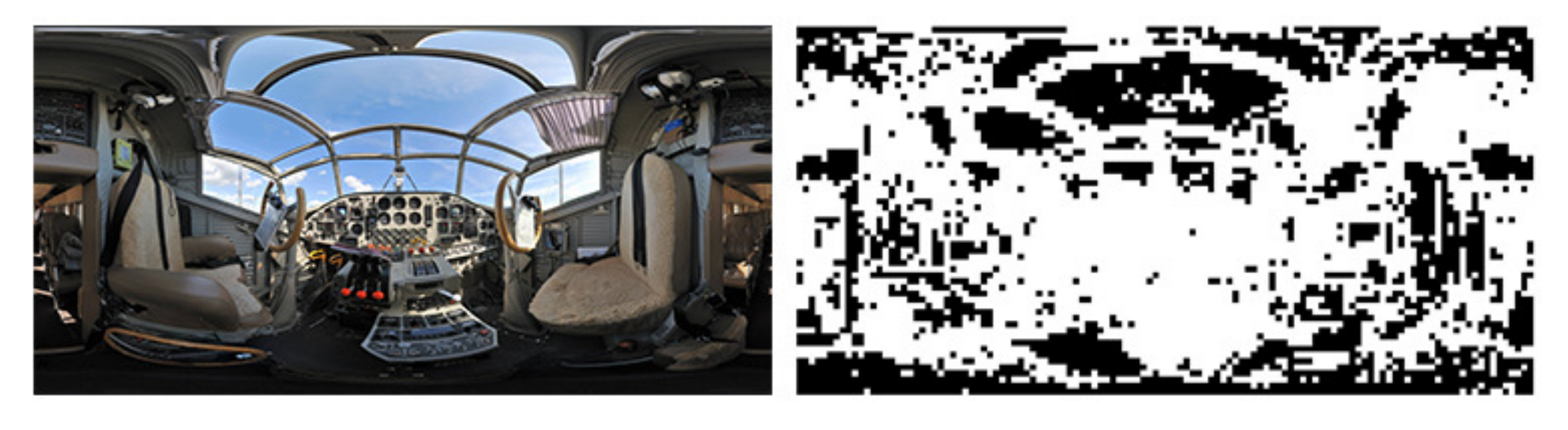}
}
\caption{An example of proposed partition mask generation: (a) The process of binary mask generation for a CTU; (b) A panoramic image and the corresponding binary mask. The CUs with small size are white, while the CUs with large size are black.}
\label{fig:cumap}
\end{figure}

\subsection{Feature Extraction Module}

During the mapping $360\degree$ videos with the ERP format, there are generally fewer contents in polar regions than in the equatorial region, so subjects frequently view the equator of the $360\degree$ frame. Meanwhile, the polar region's sampling density is significantly higher than that of the equatorial region under the ERP format. Therefore, videos with the ERP format exist problems of contents with non-uniform distribution and distortion. 
In this context, we provide a mechanism that adapts to different contents automatically and then processes smooth and textural contents separately. Specifically, a binary mask first is generated by the prior knowledge on the coding unit (CU) partition. Then, every partial convolutional layer~\cite{partialConv} guided by a CU mask activates a specified area and extracts related features. Our proposed feature extraction module uses stacked partial convolution operations. This module has a symmetrical structure, but filters' kernel sizes or shapes in every branch are adapted to both the equirectangular videos' latitudes and local contents. Details can be found as follows:

\subsubsection{Partition mask generation}

For videos via ERP, the non-uniform distribution of sampling density in the horizontal direction, partial convolutional layers focus on activating and processing smooth and textural regions, respectively, via binary masks. Thus, we exploit the CU partition information for mask generation. The reasons are three-fold: First, the CU-based mask is more accurate than the learning-based mask to distinguish smooth and textural regions, as textural regions closing to the equator are more likely to be encoded via a smaller CU, as reported in~\cite{ray2018icassp}. In contrast, smooth regions have high probabilities of being encoded via a larger CU compared to the regions with detailed texture~\cite{compressedOS}. Second, CU partition information of a compressed frame contains both geometry and content clues since the regions (\textit{e.g.}, smooth regions, and serious stretching regions) are more likely to be encoded via a large CU. Third, the CU partition information is both available on the encoder and decoder sides. There is no significant overhead on computational complexity for mask generation, which can be implemented by the parsing bitstream syntax directly~\cite{compressedOS}.

For the mask generation, we divide CU blocks in a CTU into two groups according to their depth level. Specifically, for each CTU, the pixels in CU blocks with depth level 0 and 1 are assigned with value 0, whereas the pixels in CU blocks with depth level 2 and 3 are assigned with value 1. Consequently, we obtained a binary mask map of CTU, as shown in Fig.~\ref{fig:cumap}(a). After processing all CTUs in the current frame, we obtain partition masks. As shown in Fig.~\ref{fig:cumap}(b), it is observed that the regions close to the equator contain complex local texture information (a small CU size in white), while the regions close to the pole are relatively smooth (a large CU size in black). From the equator to the poles, the distribution of CU with a large CU size is gradually increased as latitude increases.

\begin{figure}[t]
    \centering
    \includegraphics[width=0.95\linewidth]{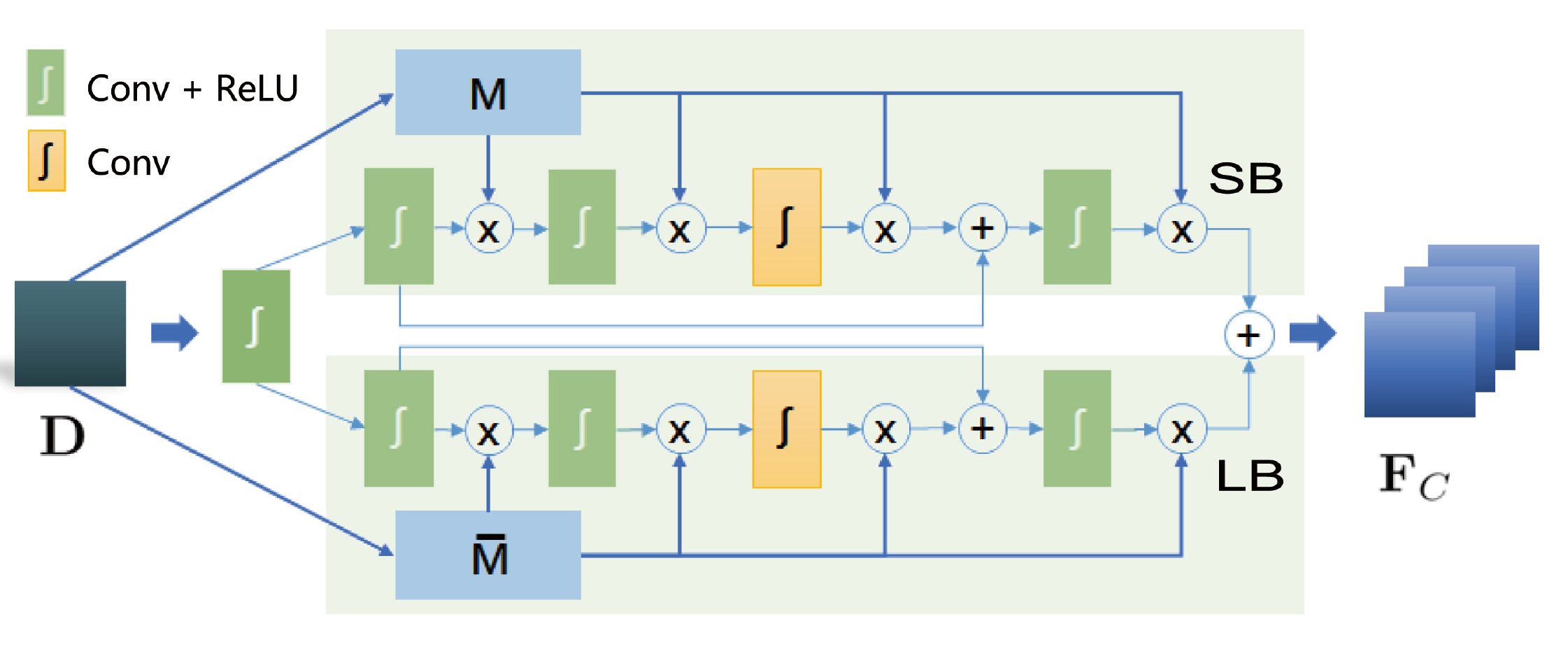}
    \caption{Proposed feature extraction module via a dual-network architecture stacked by CU-based partial convolution layers.}
    \label{fig:conAware}
\end{figure}

\subsubsection{Mask guided feature extraction}
This feature extraction module is a symmetric structure with ResNet as a backbone shown in Fig.~\ref{fig:conAware}. This symmetric structure with CU-based partial convolution can process smooth and textural regions, respectively. Specially, we regard the CU partition information-based binary mask as the following module's guiding information through the above analysis. For mathematical simplicity, the spatial resolution of the input frame $\mathbf{I}$ is denoted as $H \times W$. For each input frame, after obtaining the binary partition mask $\mathbf{M}\in \mathbb{R}^{H \times W}$, the distorted image will be fed into our proposed mask guided feature extraction module, which contains two different branches to process regions with different kernel sizes according to the mask. As shown in Fig.~\ref{fig:conAware}, the branches with a small kernel (denoted as SB) and a large kernel (denoted as LB) are a symmetrical network with the same structure, but with different parameter settings, \textit{i.e.,} kernel sizes.

We first feed the distorted image $\mathbf{I}$ into two convolution layers to obtain new feature maps $\mathbf{F}_0^B\in \mathbb{R}^{H \times W \times C_0}$, where $B \in \{$SB, LB$\}$ and $C_0$ is the channel size. Afterwards, the upper and lower branches perform partial convolutions~\cite{partialConv} under the guidance of masks $\textbf{M}^{SB}=\textbf{M}$ and $\textbf{M}^{LB}=1-\textbf{M}$, respectively. Specifically, the partial convolution makes the convolution only depend on valid pixels as
\begin{equation}
\mathbf{O}_i^{T} = \mathbf{Conv}(\mathbf{F}_i^{T} \odot \mathbf{M}^{B}),
\label{eq:ds}%
\end{equation}
where $\mathbf{F}_i^{T}$ and $\mathbf{O}_i^{T}$ are the input and output feature maps of the $i^{th}$ partial convolutional layer. And $\odot$ denotes element-wise multiplication. A residual block with three partial convolutional layers is contained in each branch, as shown in Fig.~\ref{fig:conAware}. Finally, the feature maps extracted different representing information are fused. With the mask guided partial convolution, our proposed distortion-aware feature extraction module can learn sparse feature representation, accelerating the convergence of model training.

\subsection{Feature Recalibration Module}

Videos compressed by various quality factors cause different artifacts. Although the compression quality factor is available from the decoder, it does not reflect the actual quality when the video is a transcoded or requantized video. Except the compression quality factor affected the video quality,  for videos via ERP, CUs with smooth textures closed to two polar regions have less distortion after quantization, whereas CUs with rich textures closed to the equator suffer from serious artifacts.  It can be seen that the blind system (unknown factor) is more practical in the real world. Therefore, motivated by the squeeze-and-excitation network (SENet) ~\cite{SENetPAMI}, we propose a feature recalibration module via a series of cascaded RCAB blocks to adapt to intra frames with different compression qualities. As shown in Fig.~\ref{fig:qualityAware}, each RCAB unit is composed of three parts that are a feature separation (FS) layer, an adaptive recalibration (AR) layer and a residual feature fusion (FF) layer. For mathematic simplicity, we denoted the output feature map of the ${i}^{th}$ RCAB unit as $\mathbf{X}_i$, which is also the input of the next RCAB unit. The input feature map of the first RCAB unit is $\mathbf{X}_0 = \mathbf{F}$. The details of each unit are described as follows:

\begin{figure}[t]
    \centering
    \includegraphics[scale=0.5]{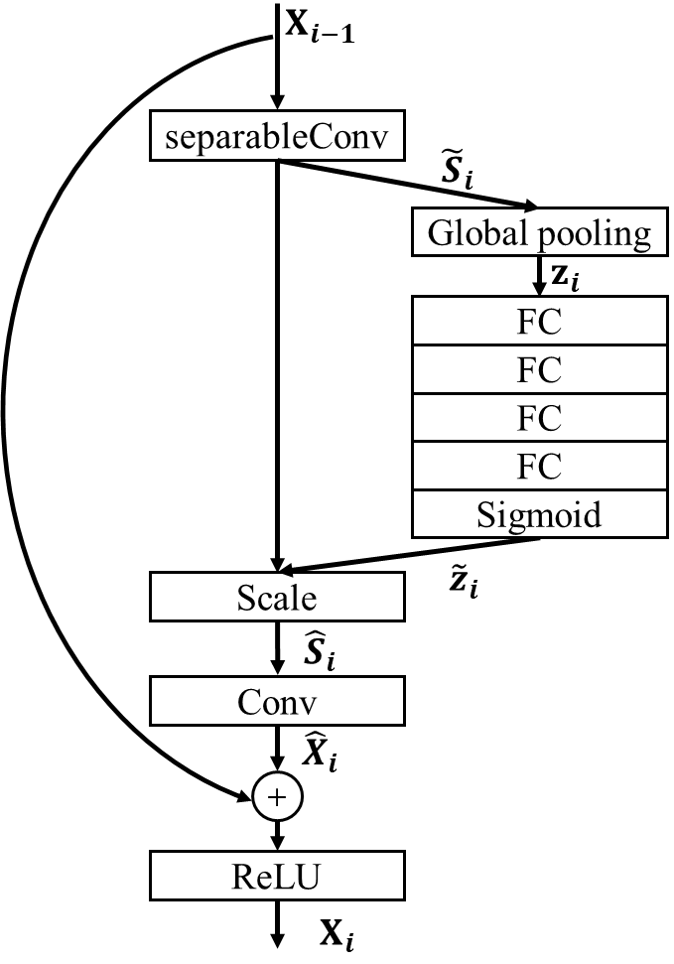}
    \caption{Illustration of proposed feature recalibration module. It is composed of feature separation (FS) layer, adaptive recalibration (AR) layer and feature fusion (FF) layer.}
    \label{fig:qualityAware}
\end{figure}

\subsubsection{Feature separation layer}
Due to the CTU level rate control, quality degradation may vary along with the spatial location. To readjust each feature map's weight along the spatial dimension, our proposed feature separation layer is utilized by performing a depthwise separable convolution~\cite{Xception}. It consists of a depthwise convolution and a pointwise convolution. The depthwise convolution performs independent spatial convolution over each channel of the input feature maps $\mathbf{X}_{i-1}$. The pointwise convolution projects the feature maps $\mathbf{S}_i \in \mathbb{R}^{H \times W \times C_i}$ output by the depthwise convolution into new feature maps $\tilde{\mathbf{S}}_i \in \mathbb{R}^{H \times W \times C_i}$.

\subsubsection{Adaptive recalibration layer}
To achieve adaptive feature maps re-weighting, we expect to explicitly learn channel interdependencies, which can increases the discriminative power of network on informative feature maps.
Similar to the method in~\cite{SENetPAMI}, the global average pooling extracts a channel-wise descriptor vector $\mathbf{z}_i \in \mathbb{R}^{C_i}$ from $\tilde{\mathbf{S}}_i$ to squeeze the global spatial information of each feature channel. After that, this channel-wise descriptor vector is projected into a high-dimensional feature space to learn channel interdependencies by performing four cascaded fully connected (FC) layers and a sigmoid layer. Denoted the adjusted channel-wise descriptor vector of feature maps $\tilde{\mathbf{S}}_i$ as $\tilde{\mathbf{z}}_i \in \mathbb{R}^{C_i}$, the recalibrated feature maps $\hat{\mathbf{S}}_i$ are obtained through the following scale operation:

\begin{equation}
\hat{\mathbf{s}}_{i}^{j} = \tilde{z}_i^j \tilde{\mathbf{s}}_{i}^{j},
\label{eq:scale}
\end{equation}
where $\hat{\mathbf{s}}_{i}^{j}$ and $\tilde{\mathbf{s}}_{i}^{j} $ are the $j^{th}$ channel of $\hat{\mathbf{S}}_i$ and $\tilde{\mathbf{S}}_i$, respectively.

\subsubsection{Residual feature fusion layer}
Since the feature maps from the distorted frame and its original version are largely similar, most values of the residual are zero. Thus, a skip connection structure is employed to learn the residual features, which makes the network easier to train. Specifically, the feature maps $\hat{\mathbf{S}}_i$ are fed into an additional convolutional layer to generate new feature maps $\tilde{\mathbf{X}}_i \in \mathbb{R}^{W \times H \times C_i}$. The output feature maps $\mathbf{X}_{i}$ of the ${i}^{th}$ RCAB is obtained:
\begin{equation}
\mathbf{X}_{i} = \mathbf{ReLU}(\tilde{\mathbf{X}}_i+\mathbf{X}_{i-1}).
\label{eq:fusion}
\end{equation}
With such attention processes from the global perspective, the feature representation of each channel is updated.

Therefore, this feature recalibration module readjusts a single feature's spatial information and features among channels to adapt to the images with various degrees of degradation.

\begin{figure}[t]
    \centering
    \includegraphics[width=0.48\textwidth]{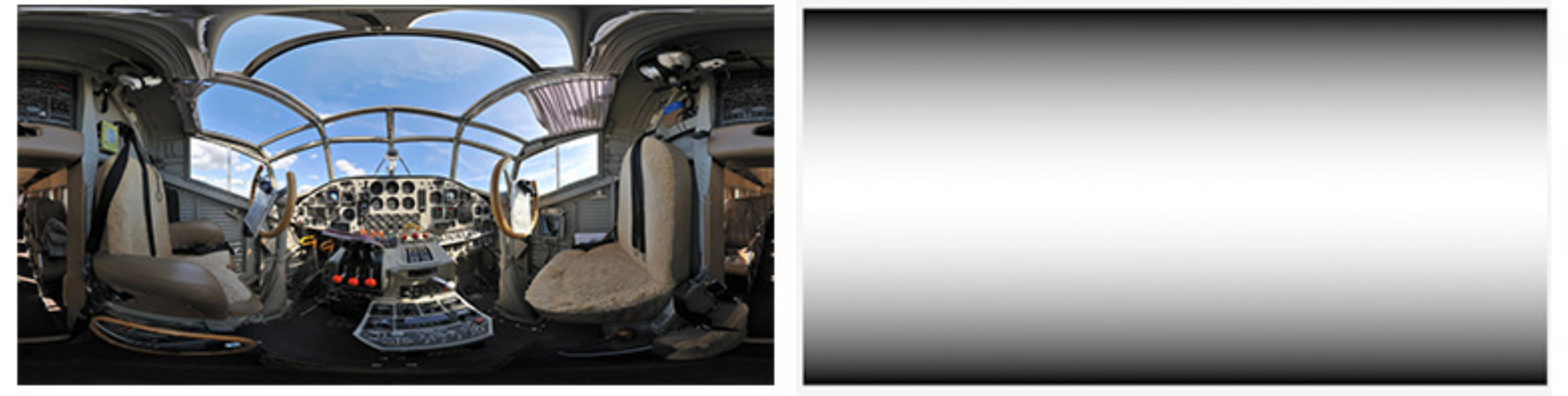}
    \caption{Visualization of a spatial weight map. The weight map is only related to the distance between the current latitude and equator, where the closer the image is to the equator, the higher the weight is.}
    \label{fig:weight}
\end{figure}

\subsection{Distortion-aware Optimization}

For image reconstruction, a residual block and a convolutional layer are performed to map the recalibrated feature maps into the restored frame $\tilde{I}\in \mathbb{R}^{H \times W}$. The residual block consists of four convolutional layers and a skip connection. Although the compression artifacts are suppressed in the feature extraction and feature recalibration modules, we still need to design a loss function to optimize the whole model. The characteristics of videos via ERP is that the human visual system prefers to high-energy regions (\textit{e.g.,} the equator and the front region of $360^{\degree}$ videos), which involve more information and are easier to attract visual attention. Thus, it can be used as prior knowledge in the optimization models of $360\degree$ videos to improve the reconstruction accuracy. In this manner, we employ MSE with a weight map that is WMSE for $360\degree$ video reconstruction.

Specifically, we assign a weight map $\mathbf{G}$ to measure the importance of contents and human attention. For the pixel at $(i,j)$, the corresponding $\mathbf{G}(i,j)$ is defined as~\cite{WMSE}.

\begin{equation}
\mathbf{G}(i,j) = \frac{g(i,j)}{\sum_{i=0}^{H-1}\sum_{j=0}^{W-1}g(i,j)},
\label{eq:wij}
\end{equation}
where $W$ and $H$ are the width and height of the $360^{\circ}$ sequence frame, respectively. And $g(i,j)$ is the scaling factor of the area, which is defined as
\begin{equation}
g(i,j) = \cos ((j-\frac{H}{2}+\frac{1}{2})\cdot \frac{\pi}{H}).
\label{eq:wij}
\end{equation}
The weight map is visualized in Fig.~\ref{fig:weight}, where the important area closing to the equator has a higher weight, while as it gets closer to the poles, the weight decreases. Combining the weight map, WMSE can be defined as
\begin{equation}
\mathcal{L}_{re}(\mathbf{P}) = \frac{1}{N}\sum_{\mathbf{p}\in\mathbf{P}}^{N}(\mathbf{Y}(\mathbf{p})- \tilde{\mathbf{I}}(\mathbf{p}))^2 \odot \mathbf{G}(\mathbf{p}),
\label{eq:Iffd}
\end{equation}
where $\mathbf{p}$ is the index of the pixel and $\mathbf{P}$ is the 3D tensor with size $N=B\times H_p \times W_p$. $B$ is the batch size. $H_p$ and $W_p$ are the width and height of each patch.

For the perceptual approach on $360\degree$ videos, we evaluate the difference of the viewport images between reconstruction images and groundtruth images, which is marked as $\mathcal{L}_{v}$. Specifically, we describe one of viewports $\left(\mathbf{p}_{\mathrm{v}}=(\phi_{v}, \theta_{v}\right)$ how to project to the 2D plane $\mathbf{C}=(x, y)$ under the unit sphere, which needs the whole set of transformations applied across different coordinate systems \cite{li2019viewport, snyder1987map}. Specially, the coordinates of a point $p$ on the viewport ($x$, $y$) ($x$ $\in$[1, $W_v$], $y$ $\in$[1, $H_v$]) in $\mathbf{C}$ are transformed to a longitude coordinate and a latitude coordinate in the spherical coordinate system $p_{s}=(\theta_{x,y}, \phi_{x,y})$. We denote the width and height of the viewport as W and H. Firstly, the pixel location $(x, y)$ is converted to $\left(f_{x}, f_{y}\right),$ which has the same scale as the unit sphere:
\begin{equation}
f_{x}=\frac{2 x-1-W_v}{W_v} \cdot \tan \frac{\alpha_{W}}{2},
\end{equation}
\begin{equation}
f_{y}=-\frac{2 y-1-H_v}{H_v} \cdot \tan \frac{\alpha_{H}}{2},
\end{equation}
where $\alpha_{W}$ and $\alpha_{H}$ are the angular ranges of the viewport, corresponding to $W_v$ and $H_v$, respectively. Then, the corresponding spherical location $\left(\phi_{x, y}, \theta_{x, y}\right)$ is obtained by
\begin{equation}
\phi_{x, y}=\phi_{v}+\arctan \left(\frac{f_{x} \sin c}{\rho \cos \theta_{v} \cos c-f_{y} \sin \theta_{v} \sin c}\right),
\end{equation}

\begin{equation}
\theta_{x, y}=\quad \arcsin \left(\cos c \sin \theta_{v}+\frac{f_{y} \sin c \cos \theta_{v}}{\rho}\right),
\end{equation}
where

\begin{equation}
\rho=\sqrt{f_{x}^{2}+f_{y}^{2}}, \quad c=\arctan \rho
\end{equation}

Next, to ERP, the corresponding pixel locations $\left(p_{x, y}, q_{x, y}\right)$ of $\left(\phi_{x, y}, \theta_{x, y}\right)$ can be obtained by:

\begin{equation}
p_{x, y}=\left(\frac{\phi_{x, y}}{360^{\circ}}+\frac{1}{2}\right) W +\frac{1}{2},
\end{equation}

\begin{equation}
q_{x, y}=\left(\frac{1}{2}-\frac{\theta_{x, y}}{180^{\circ}}\right) H +\frac{1}{2}.
\end{equation}
Finally, the pixel values at $(x, y)$ in $\mathbf{C}$ are obtained by bilinear interpolation at $\left(p_{x, y}, q_{x, y}\right)$, generating the $2 \mathrm{D}$ content of viewports. Thus, we obtain viewport images from the reconstruction image ($\mathbf{V}$) and the groundtruth image ($\tilde{\mathbf{V}}$). The $\mathcal{L}_{v}$ can be defined by MSE of $\mathbf{V}$ and $\tilde{\mathbf{V}}$:

\begin{equation}
\mathcal{L}_{v}(\mathbf{P}) = \frac{1}{N}\sum_{\mathbf{p}\in\mathbf{P}}^{N}\sqrt{(\mathbf{V}(\mathbf{p})- \tilde{\mathbf{V}}(\mathbf{p}))^2},
\label{eq:Iffd}
\end{equation}
where $\mathbf{p}$ is the index of the pixel and $\mathbf{P}$ is the 3D tensor with size $N=B\times H_p \times W_p$. $B$ is the batch size. This perceptual approach is generally of better performance but with the risk of overfitting, so the above two types of approaches combined are controlled by a weight factor ($\lambda$). Finally, the loss function for training is a combination of $\mathcal{L}_{re}$ and $\mathcal{L}_{v}$. Empirically, $\lambda$ is assigned with value 0.5.

\begin{equation}
\mathcal{L}= \mathcal{L}_{re}+\lambda\mathcal{L}_{v}.
\end{equation}

\subsection{Integration in the HEVC Framework}

\begin{figure}[t]
    \centering
    \includegraphics[width=0.5\textwidth]{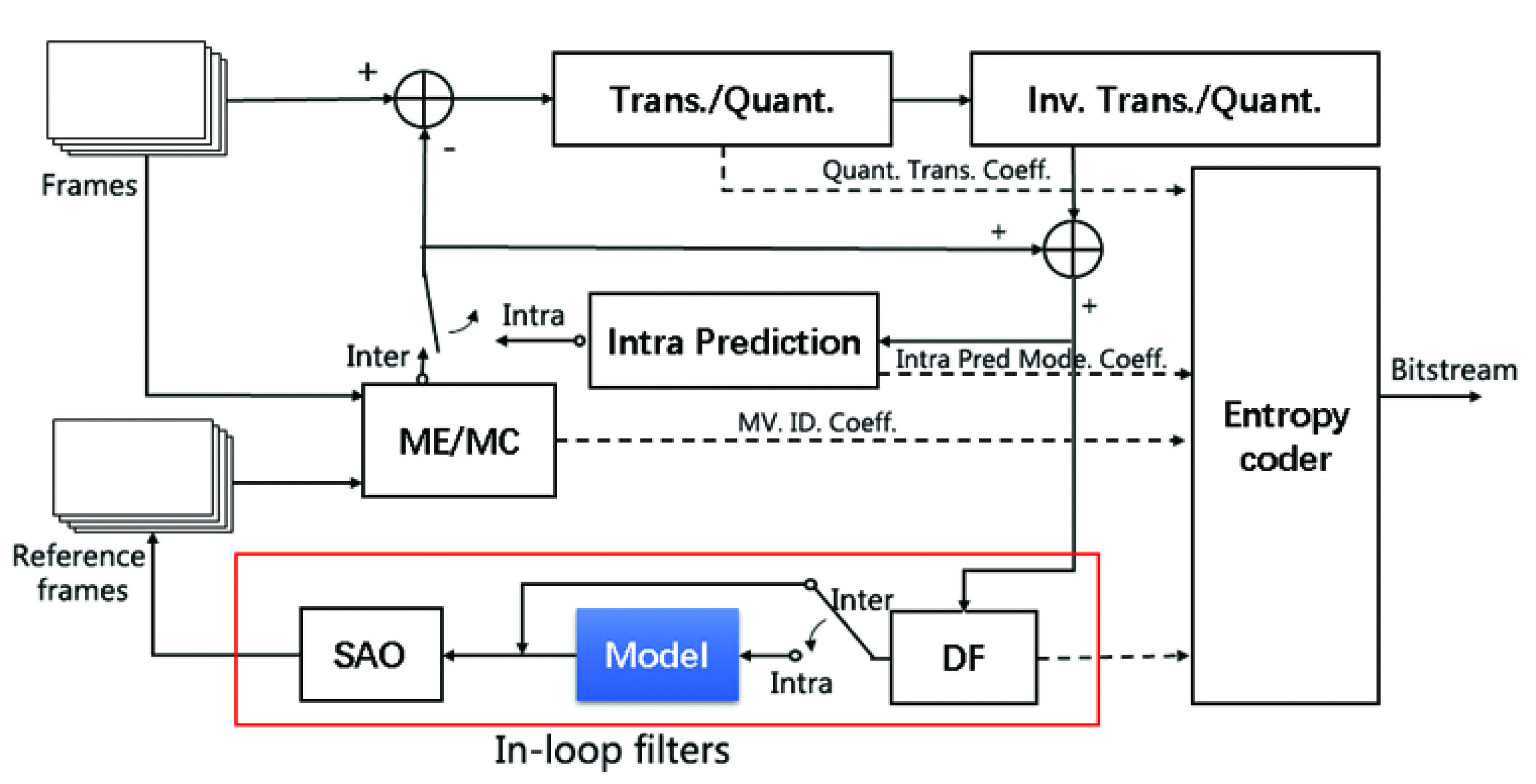}
   \caption{Our proposed model is integrated into HEVC as an additional filter between DF and SAO modules.}
    \label{fig:HEVC_framework}
\end{figure}

We integrate the trained model in the HEVC codec as in~\cite{dlvc}.
As shown in Fig.~\ref{fig:HEVC_framework}, our proposed model is applied as an additional filter between the DF and SAO module. Specifically, each reconstructed CTU of the intra frame is filtered by DF firstly before fed into our model. After DF, our proposed model can enhance the quality of reconstructed CTUs of the intra frame. Then, the SAO is employed further to reduce artifacts in the outputs of our model adaptively. As the signal reconstruction quality improved, the overhead bits for an offset signal of SAO may be reduced. Notably, our proposed method is only enabled for the intra frames.

\begin{table*}[t]
  \centering
  \caption{The results of RHCNN, SEFCNN and Ours compared with HM under configuration of AI for BD-rate (\%) of  PSNR, WSPSNR, PSNR of viewport 0 and viewport 1 for 360$^o$ video sequences (HM coding as anchor).}
    \begin{tabular}{|c|c|c|c|c|c|c|c|c|c|c|c|c|}
    \hline
    \multicolumn{1}{|c|}{\multirow{2}[4]{*}{AI}} & \multicolumn{3}{c|}{PSNR} & \multicolumn{3}{c|}{WSPSNR} & \multicolumn{3}{c|}{PSNR(VP0)} & \multicolumn{3}{c|}{PSNR (VP1)} \bigstrut\\
\cline{2-13}          & RHCNN & SEFCNN & Ours  & RHCNN & SEFCNN & Ours  & RHCNN & SEFCNN & Ours  & RHCNN & SEFCNN & Ours \bigstrut\\
    \hline
    4K    & -6.0\% & -6.9\% & \textbf{-8.9\%} & -6.3\% & -7.1\% & \textbf{-8.9\%} & -4.8\% & -5.5\% & \textbf{-7.4\%} & -5.1\% & -5.8\% & \textbf{-7.7\%} \bigstrut\\
    \hline
    6K    & -4.9\% & -6.5\% & \textbf{-9.0\%} & -5.4\% & -6.7\% & \textbf{-8.8\%} & -4.3\% & -5.1\% & \textbf{-6.9\%} & -2.4\% & -4.0\% & \textbf{-7.0\%} \bigstrut\\
    \hline
    8K    & -4.4\% & -6.1\% & \textbf{-9.0\%} & -5.1\% & -6.6\% & \textbf{-9.2\%} & -3.6\% & -4.6\% & \textbf{-7.0\%} & -3.8\% & -4.9\% & \textbf{-7.5\%} \bigstrut\\
    \hline
    Average & -5.1\% & -6.5\% & \textbf{-8.9\%} & -5.6\% & -6.8\% & \textbf{-9.0\%} & -4.3\% & -5.0\% & \textbf{-7.1\%} & -3.8\% & -4.9\% & \textbf{-7.4\%} \bigstrut\\
    \hline
    \multicolumn{1}{r}{} & \multicolumn{1}{r}{} & \multicolumn{1}{r}{} & \multicolumn{1}{r}{} & \multicolumn{1}{r}{} & \multicolumn{1}{r}{} & \multicolumn{1}{r}{} & \multicolumn{1}{r}{} & \multicolumn{1}{r}{} & \multicolumn{1}{r}{} & \multicolumn{1}{r}{} & \multicolumn{1}{r}{} & \multicolumn{1}{r}{} \bigstrut\\
    \hline
    \multicolumn{1}{|c|}{\multirow{2}[4]{*}{LDB}} & \multicolumn{3}{c|}{PSNR} & \multicolumn{3}{c|}{WSPSNR} & \multicolumn{3}{c|}{PSNR(VP0)} & \multicolumn{3}{c|}{PSNR(VP1)} \bigstrut\\
\cline{2-13}          & RHCNN & SEFCNN & Ours  & RHCNN & SEFCNN & Ours  & RHCNN & SEFCNN & Ours  & RHCNN & SEFCNN & Ours \bigstrut\\
    \hline
    4K    & -4.4\% & -4.9\% & \textbf{-5.2\%} & -4.4\% & -4.8\% & \textbf{-5.2\%} & -4.3\% & -4.6\% & \textbf{-5.1\%} & -3.6\% & \textbf{-4.2\%} & \textbf{-4.2\%} \bigstrut\\
    \hline
    6K    & -1.5\% & -1.7\% & \textbf{-1.8\%} & -1.4\% & -1.6\% & \textbf{-1.8\%} & -1.2\% & -1.4\% & \textbf{-1.5\%} & -1.0\% & -1.1\% & \textbf{-1.4\%} \bigstrut\\
    \hline
    8K    & -5.8\% & -6.5\% & \textbf{-7.1\%} & -5.9\% & -6.4\% & \textbf{-7.1\%} & -3.2\% & -3.6\% & \textbf{-4.2\%} & -4.8\% & -5.3\% & \textbf{-6.2\%} \bigstrut\\
    \hline
    Average & -3.9\% & -4.3\% & \textbf{-4.7\%} & -3.9\% & -4.3\% & \textbf{-4.7\%} & -2.9\% & -3.2\% & \textbf{-3.6\%} & -3.2\% & -3.5\% & \textbf{-4.0\%} \bigstrut\\
    \hline
    \multicolumn{1}{r}{} & \multicolumn{1}{r}{} & \multicolumn{1}{r}{} & \multicolumn{1}{r}{} & \multicolumn{1}{r}{} & \multicolumn{1}{r}{} & \multicolumn{1}{r}{} & \multicolumn{1}{r}{} & \multicolumn{1}{r}{} & \multicolumn{1}{r}{} & \multicolumn{1}{r}{} & \multicolumn{1}{r}{} & \multicolumn{1}{r}{} \bigstrut\\
    \hline
   \multicolumn{1}{|c|}{\multirow{2}[4]{*}{LDP}} & \multicolumn{3}{c|}{PSNR} & \multicolumn{3}{c|}{WSPSNR} & \multicolumn{3}{c|}{PSNR(VP0)} & \multicolumn{3}{c|}{PSNR(VP1)} \bigstrut\\
\cline{2-13}          & RHCNN & SEFCNN & Ours  & RHCNN & SEFCNN & Ours  & RHCNN & SEFCNN & Ours  & RHCNN & SEFCNN & Ours \bigstrut\\
    \hline
    4K    & -4.8\% & -5.2\% & \textbf{-5.5\%} & -4.6\% & -5.0\% & \textbf{-5.3\%} & -4.2\% & -4.5\% & \textbf{-4.9\%} & -3.9\% & -4.2\% & \textbf{-4.4\%} \bigstrut\\
    \hline
    6K    & -1.4\% & \textbf{-1.7\%} & \textbf{-1.7\%} & -1.4\% & \textbf{-1.7\%} & -1.6\% & -1.1\% & \textbf{-1.3\%} & -1.2\% & -0.9\% & -1.1\% & \textbf{-1.2\%} \bigstrut\\
    \hline
    8K    & -5.9\% & -6.5\% & \textbf{-7.1\%} & -5.8\% & -6.4\% & \textbf{-7.1\%} & -3.1\% & -3.4\% & \textbf{-4.1\%} & -4.7\% & -5.2\% & \textbf{-6.1\%} \bigstrut\\
    \hline
    Average & -4.0\% & -4.5\% & \textbf{-4.8\%} & -4.0\% & -4.4\% & \textbf{-4.7\%} & -2.8\% & -3.1\% & \textbf{-3.4\%} & -3.1\% & -3.5\% & \textbf{-3.9\%} \bigstrut\\
    \hline
    \multicolumn{1}{r}{} & \multicolumn{1}{r}{} & \multicolumn{1}{r}{} & \multicolumn{1}{r}{} & \multicolumn{1}{r}{} & \multicolumn{1}{r}{} & \multicolumn{1}{r}{} & \multicolumn{1}{r}{} & \multicolumn{1}{r}{} & \multicolumn{1}{r}{} & \multicolumn{1}{r}{} & \multicolumn{1}{r}{} & \multicolumn{1}{r}{} \bigstrut\\
    \hline
    \multicolumn{1}{|c|}{\multirow{2}[4]{*}{RA}} & \multicolumn{3}{c|}{PSNR} & \multicolumn{3}{c|}{WSPSNR} & \multicolumn{3}{c|}{PSNR(VP0)} & \multicolumn{3}{c|}{PSNR(VP1)} \bigstrut\\
\cline{2-13}          & RHCNN & SEFCNN & Ours  & RHCNN & SEFCNN & Ours  & RHCNN & SEFCNN & Ours  & RHCNN & SEFCNN & Ours \bigstrut\\
    \hline
    4K    & -3.5\% & -5.8\% & \textbf{-6.2\%} & -3.3\% & -5.7\% & \textbf{-6.2\%} & -2.6\% & -5.6\% & \textbf{-6.2\%} & -2.2\% & -4.6\% & \textbf{-4.9\%} \bigstrut\\
    \hline
    6K    & -4.4\% & -5.1\% & \textbf{-5.2\%} & -4.2\% & -4.9\% & \textbf{-5.0\%} & -5.1\% & \textbf{-5.8\%} & \textbf{-5.8\%} & -0.8\% & -1.3\% & \textbf{-1.5\%} \bigstrut\\
    \hline
    8K    & -7.1\% & -9.6\% & \textbf{-10.4\%} & -7.2\% & -9.5\% & \textbf{-10.4\%} & -5.6\% & -7.5\% & \textbf{-8.1\%} & -5.8\% & -8.2\% & \textbf{-9.2\%} \bigstrut\\
    \hline
    Average & -5.0\% & -6.8\% & \textbf{-7.2\%} & -4.9\% & -6.7\% & \textbf{-7.2\%} & -4.4\% & -6.3\% & \textbf{-6.7\%} & -2.9\% & -4.7\% & \textbf{-5.2\%} \bigstrut\\
    \hline
    \end{tabular}%
  \label{tab:AI_LDB_LDP_RA}%
\end{table*}%

\section{Experimental Results}
\label{sec:exp}
In this section, we conduct extensive experiments to evaluate the performance of our proposed model. We first describe the experimental settings such as training sets, network training protocols, and test configurations in the following subsections. Details on the performance comparison between the proposed model and the state-of-the-art models are described as follows.

\subsection{Experimental Settings}
\subsubsection{Training set}
To guarantee the diversity of scenes, we generate training sets by selecting representative scenes from the SUN360 panorama dataset~\cite{xiao2012recognizing}. It contains 83 different categories with image resolution $9104\times 4552$, including both indoor and outdoor scenes. We rando mly select 1 or 2 representative scenes for each category, resulting in 130 images in total. For data augmentation, all pristine images are resized into three different resolutions, including $3840 \times 1920$ (denoted as 4K), $6144 \times 3072$ (denoted as 6K) and $8192 \times 4096$ (denoted as 8K), to improve the generality of the proposed loop filtering model capable of handling of various resolutions.

To simulate the compression distortion in 360$^o$ videos, the color space of all images are first converted from RGB to YUV and then compressed by the HEVC reference test model HM (16.18) + 360Lib. Specifically, all images are encoded under all intra configuration with the following QP settings where QP $\in \{27, 32, 37, 42\}$. The video compressed by low QP like 22 has high reconstruction quality, whereas loop filtering schemes are more suitable for videos with low reconstruction quality compressed by high QPs, especially for panoramic videos with higher resolution. We encode videos with a 42 QP value for performance evaluation. The encoder is modified to output the CU partition information for generating the partition mask.
Since we focus on the restoration of the luminance component of intra frames, the training set is built by extracting patches from the Y-component of decompressed images. For each distorted image, we randomly sample 50 non-overlapped patches with size $64\times 64$ from the middle area of the uncompressed frame, the distorted frame, the corresponding partial mask and the weight map. In contrast, we sample 30 patches close to the two poles. Due to the ERP format of 360$\degree$ video, the content of polar region is usually less than that of equatorial region, and viewers often watch the central region. Finally, patches with different QPs and resolutions are mixed and shuffled for the model training.

\begin{figure*}[t]

\centering
\includegraphics[width=\linewidth]{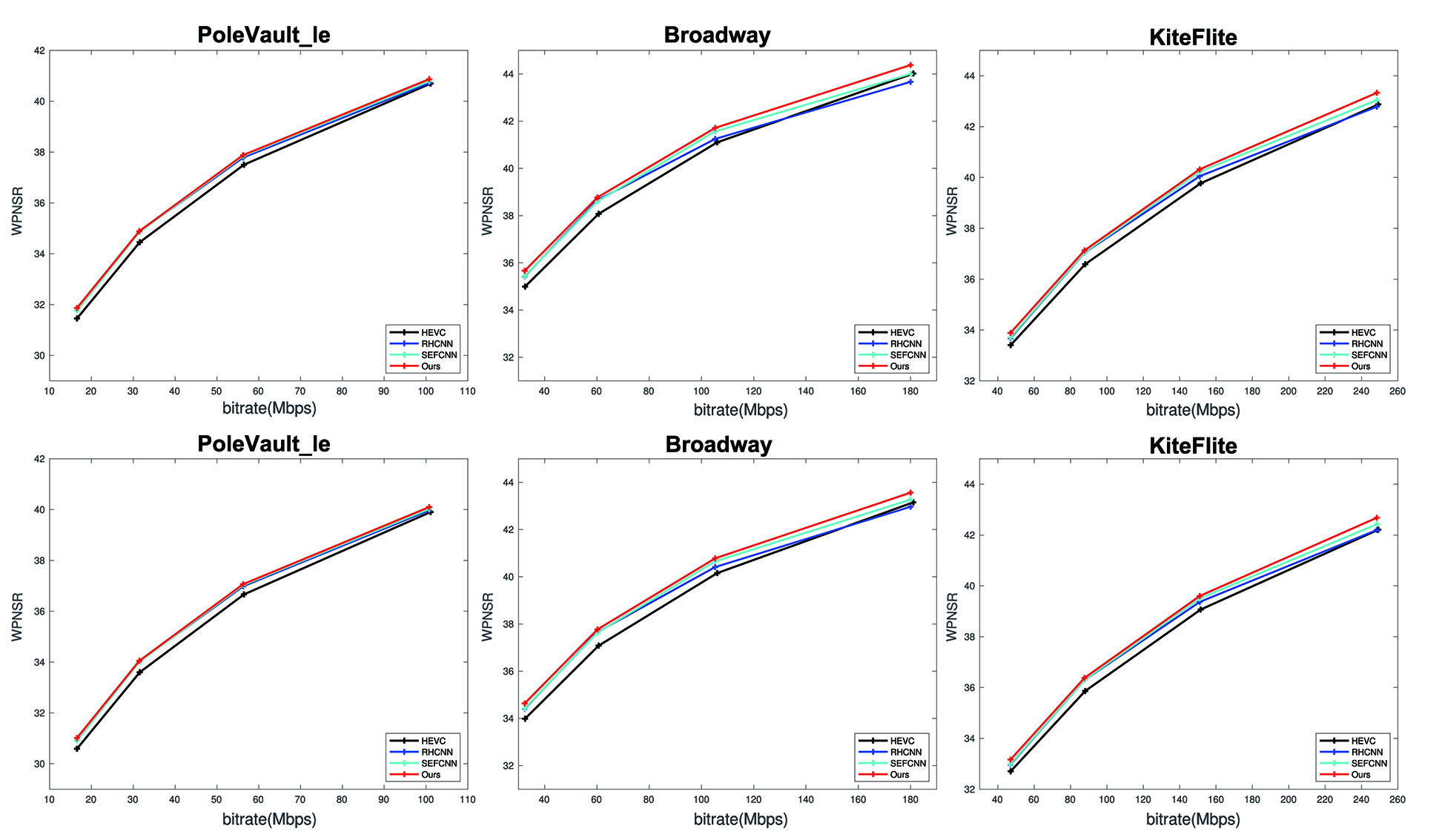}

\caption{Typical R-D curves in terms of PSNR and WPSNR for different models evaluated on video sequences under AI configuration. From the left column to the right column: ``PoleVault\_le''(4K), ``Broadway''(6K), ``KiteFlite''(8K).}
\label{fig:rdcurves}
\end{figure*}

\subsubsection{Parameter settings}
By default, the channel size of all convolutional layers are set as 128. The kernel sizes of convolutional layers in the feature extraction module are set as $3 \times 3$ and $5 \times 5$ for the SB branch and the LB branch, respectively. The feature recalibration module contains four RCABs. The four FC layers of adaptive recalibration layer are with 512, 256, 128 and 128 nodes in each RCAB, respectively. Due to the limitation of the patch size, we reduce the FOV angle to 5 degrees, so $\alpha_W$ and $\alpha_H$ are 5 degrees. To speed up the converging process and achieve global optimal parameters, Xavier initialization and adaptive moment estimation (Adam) are employed as the gradient descent optimization algorithm in model training. The learning rate with the initial value of 0.0001 decreases as the iteration number increases.

\subsubsection{Test configurations}
To verify the feasibility and versatility of our proposed model, we chose 11 video sequences~\cite{CTC} for testing, where the scenes are completely different from those of the training set. The test video resolution varies from 4K to 8K. There are four 4K videos (``AerialCity", ``DrivingInCity", ``DrivingInCountry" and ``PoleVault"), four 6K videos (``Balboa", ``BranCastle2", ``Broadway" and ``Landing2") and three 8K videos (``Harbor", ``KiteFlite" and ``Trolley"). The weighted-to-spherically-uniform PSNR (WS-PSNR)~\cite{WMSE} is employed to measure the distortion of 360$^o$ content in the observation space.

To evaluate the performance of the proposed method, experiments are conducted under four configurations, including All Intra (AI), Low-delay B (LDB), Low-delay P (LDP), and Random Access (RA). All settings follow the common test conditions (CTC)~\cite{boyce2016common}, except one of the QP settings and the number of the coding frame. As for AI, the first 50 successive frames of each test sequences are encoded for performance comparison. The frame numbers of each sequence for LDB, LDP and RA configurations are 50, 50 and 64, respectively. In this paper, the CNN based loop filtering model is enabled only for intra frames to measure its contributions to the R-D performance as reference frames. Meanwhile, we evaluate the PSNR of videos from viewpoint 0 and 1 (V0 and V1).

\begin{figure*}[t]
\subfigure[]{
\centering
\includegraphics[width=0.98\linewidth]{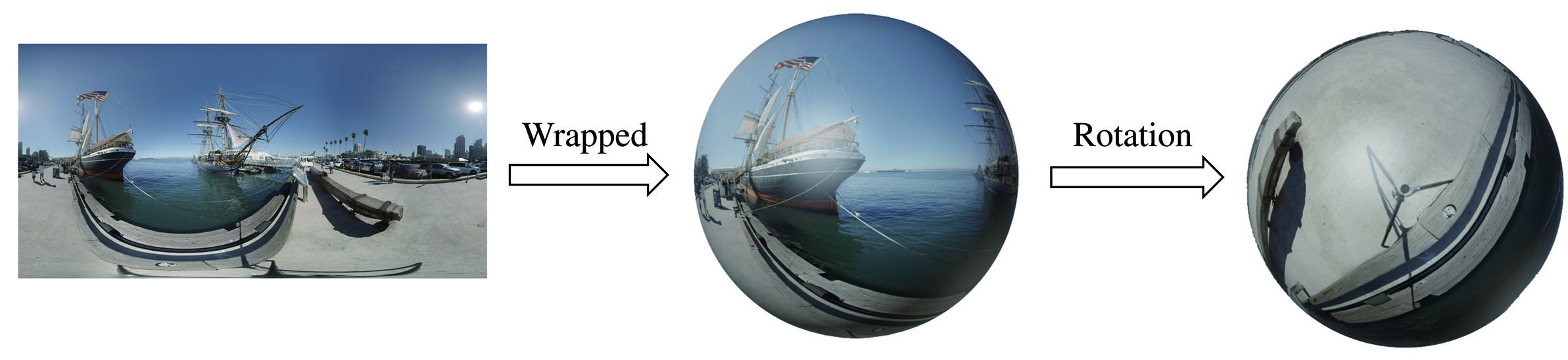}
}
\subfigure[]{
\centering
\includegraphics[width=0.98\linewidth]{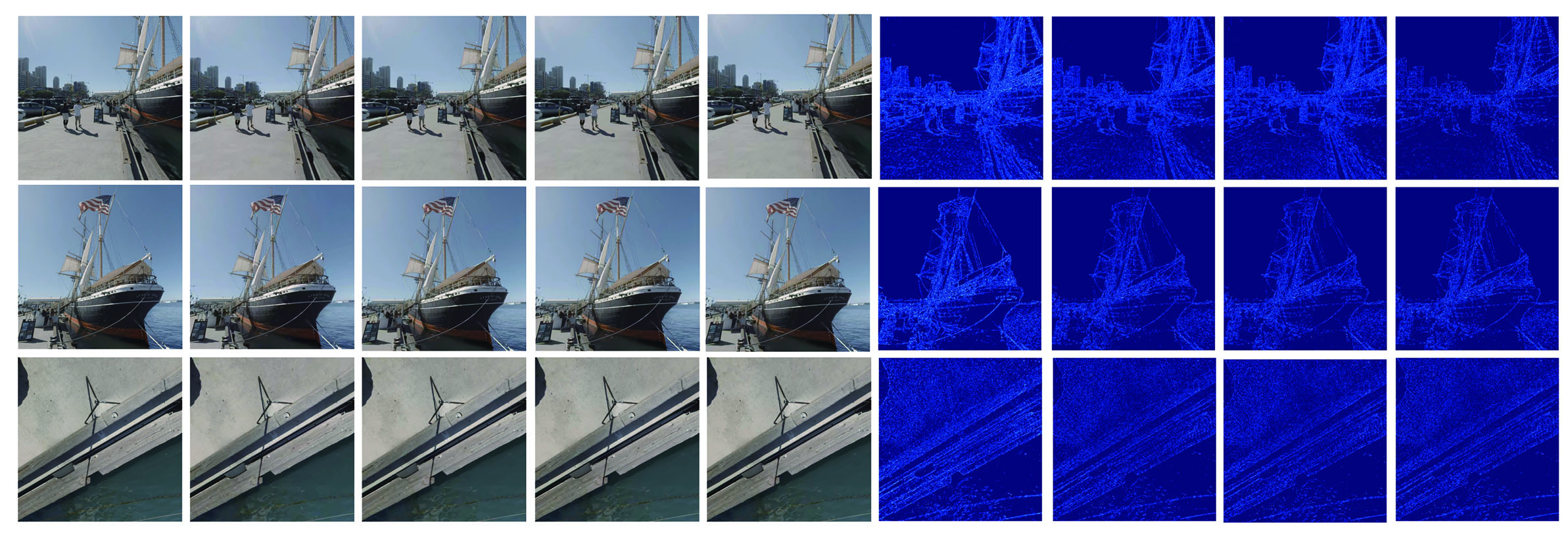}
}
\caption{Example of the frame for FoV images quality comparison. (a) Illustration of FoV extraction generation. From left to right: The ﬁrst frame of the video sequence Harbor (8K), its sphere representation version, FoV images on the sphere; (b) FoV images from different view directions. From left to right: uncompressed, compressed by HEVC, restored by RHCNN, SEFCNN and Ours. The last four images are the differences of images between the compressed image and the restored FOV images via RHCNN, SEFCNN and Ours comparing with the ground truth, respectively.}
\label{fig:fov}
\end{figure*}

\subsection{Comparisons with the State-of-the-Art Methods}
In this subsection, the proposed model is compared with RHCNN~\cite{zhang2018residual} and SEFCNN \cite{ding2019switchable}. HM 16.18 is utilized as the anchor. RHCNN and SEFCNN are also embedded into HM between the DF and SAO modules. Since the RHCNN is not adaptive for different compression artifacts, we train it on the training sets with four different QP settings. For each configuration, QPs are set as $\{27, 32, 37, 42\}$ to test the performance from high to low bitrates. For SEFCNN, it trains a model for these four QPs.

\subsubsection{Overall performance}

The comparison results of loop filtering models under the AI, LDB, LDP and RA configurations are summarized in Table~\ref{tab:AI_LDB_LDP_RA}. The negative values of BDBR represent the ratio of bitrate saving compared with the anchor HM. It is observed that our proposed model holds significant superiority over HM, RHCNN and SEFCNN under AI configuration.  In this context, we compared models on BDBR in terms of PSNR, WSPSNR from the whole EPR videos and the two viewport videos (V0 and V1). The locations and angles of viewports follow the CTC. In particular, our proposed model can achieve 3.8\% and 3.4\% bit rate reduction in terms of PSNR and WPSNR compared with RHCNN, and 2.4\% and 2.2\% bit rate reduction in terms of PSNR and WPSNR compared with SEFCNN under AI configuration. For LDB, LDP and RA, we only improve I frames to evaluate influence on B and P frames since they reference I frames. Thus, compared with improving all frames, the improvement of this method will be relatively small. To illustrate the performance of our proposed model, some typical rate-distortion (R-D) curves of all models evaluated on 360$^o$ video sequences under AI configuration are also provided in Fig.~\ref{fig:rdcurves}. It is observed that our proposed model outperforms other methods by a significant margin for sequences with different resolutions from low bitrates to high bitrates.

\subsubsection{Local FoV image quality comparison}
When enjoying a 360$^o$ video, a viewer wears an HMD where the 360$^o$ video will be unwrapped onto the sphere. Users can choose FoVs from top to down ($\phi \in [-\pi/2, \pi/2]$) and left to right ($\theta \in [-\pi, \pi]$). From Table 1, It is observed that RHCNN, SEFCNN and our model can improve the image quality in terms of global PSNR and WPSNR values. However, we found RHCNN and SEFCNN without $\mathcal{L}_{v}$ have poor performance on the viewport images. To better show our proposed model's superiority intuitively, we take the first frame of ``Harbor'' under AI configuration with QP=37 as shown in Figure \ref{fig:fov}, and the ERP image is projected to FOV images~\cite{li2019viewport, snyder1987map}. The field angle is $75^o$. The FOV image resolution of an 8K image (``Harbor'') is $1816 \times 1816$ from three view directions. From left to right in Fig.~\ref{fig:fov} (a), it illustrates the ﬁrst frame of the video sequence Harbor (8K), its sphere representation version and FoV images on the sphere; and Fig.~\ref{fig:fov} (b) shows FoV images from different view directions. These are uncompressed and compressed images and images restored by RHCNN, SEFCNN and Ours from left to right. For better visualization, we give the difference of image between the compressed FOV image via HEVC and the ground truth, and the differences of the restored FOV images via RHCNN, SEFCNN and Ours, comparing with the ground truth, respectively. Our proposed model can recover structural information and suppress noises around the edge to improve the quality of FoV images.

\begin{table}[t]
  \centering
  \caption{Performance comparison of using a single model for QP datasets not contained in the training set in terms of average WS-PSNR (dB)}
    \begin{tabular}{c|c|c|c|c}
    \hline
    \hline
    \textbf{QP} & \textbf{Methods} & \textbf{4K} & \textbf{6K} & \textbf{8K} \bigstrut\\
    \hline
    \hline
    \multirow{3}[6]{*}{\textbf{25}} & \textbf{HM} & 43.17  & 45.18  &  45.67     \bigstrut\\
\cline{2-5}          & \textbf{RHCNN}             & 43.02  & 44.91  &  45.44     \bigstrut\\
\cline{2-5}          & \textbf{SEFCNN}            & 43.08  & 45.11  &  45.71     \bigstrut\\
\cline{2-5}          & \textbf{Ours}              & \textbf{43.25}  & \textbf{45.43}  &  \textbf{46.11}  \bigstrut\\
    \hline
    \hline
    \multirow{3}[6]{*}{\textbf{30}} & \textbf{HEVC} & 40.40       &   42.04      &   42.73  \bigstrut\\
\cline{2-5}          & \textbf{RHCNN}               & 40.54       &   42.14      &   42.90   \bigstrut\\
\cline{2-5}          & \textbf{SEFCNN}              & 40.64       &   42.35      &   43.13   \bigstrut\\
\cline{2-5}          & \textbf{Ours}                & \textbf{40.75}       &   \textbf{42.50}      &   \textbf{43.27}    \bigstrut\\
    \hline
    \hline
    \multirow{3}[6]{*}{\textbf{35}} & \textbf{HM} & 37.59   & 38.95    & 39.62    \bigstrut\\
\cline{2-5}          & \textbf{RHCNN}             & 37.99   & 39.41    & 40.13    \bigstrut\\
\cline{2-5}          & \textbf{SEFCNN}            & 37.97   & 39.35    & 40.06    \bigstrut\\
\cline{2-5}          & \textbf{Ours}              & \textbf{38.07}   & \textbf{39.49}   & \textbf{40.19}   \bigstrut\\ 
    \hline
    \hline
    \multirow{3}[6]{*}{\textbf{40}} & \textbf{HM} & 34.85  & 36.02  & 36.61  \bigstrut\\
\cline{2-5}          & \textbf{RHCNN}             & 35.21  & 36.41  & 36.95  \bigstrut\\
\cline{2-5}          & \textbf{SEFCNN}            & 35.21  & 36.40  & 36.95   \bigstrut\\
\cline{2-5}          & \textbf{Ours}              & \textbf{35.29}  & \textbf{36.48}  & \textbf{37.05}  \bigstrut\\
    \hline
    \hline
    \end{tabular}%
  \label{tab:Test_in_different_QP1}%
\end{table}%

\subsubsection{Verification of training a single model for multiple QPs}
Our proposed model is a single model trained on the image patches with mixed QP settings. To verify the applicability of our proposed model, we evaluate the models on the testing dataset with different QP settings that are not included in the training set. The model RHCNN needs to be trained multiple times for each QP band, so we choose the model trained on the dataset compressed under the same QP. The results are shown in Table~\ref{tab:Test_in_different_QP1}. It is observed that our proposed model can still achieve superior performance in terms of WPSNR.

\subsection{Runtime analysis}
The encoding and decoding runtimes are evaluated in this sub-section. The testing hardware device is a server with Intel Xeon CPU E5-2690 v4, 256GB memory and NVIDIA Tesla P100 GPU (16GB memory). In addition, the HM + 360Lib and TensorFlow are compiled by GCC 5.4.0 in the Ubuntu16.04. The testing results are shown in Table \ref{tab:times} where it provides the average encoder and decoder running times (seconds/frame). When evaluating the coding complexity, the $\triangle$T is calculated as \cite{jia2019content},
\begin{equation}
{\triangle T = \frac{T^{'}- T}{T}}
\label{eq:fusion}
\end{equation}
where $T$ is the original running time of the HM + 360Lib reference software, and $T'$ is the running time of our proposed model. The numerous convolutional operations in our model require much time; the running time of the encoder is tolerable. However, for the decoder, this process requires a great deal of time. Due to the limitation of GPU memory, it is difficult to reconstruct the whole frame at once, especially for 8K resolution, so we need to crop each frame into small patches with 1024 $\times$ 1024, process them one by one, and then merge these patches after restoration. Moreover, every patch overlaps 20 pixels to ensure the boundary areas better processing. Thus, it takes a lot of time to process these patches sequentially. If the number of GPU and memories are not limited, multiple patches can be processed in parallel. Such that the encode and decode speed will be significantly improved.

\begin{figure}[t]
    \centering
    \includegraphics[width=0.48\textwidth]{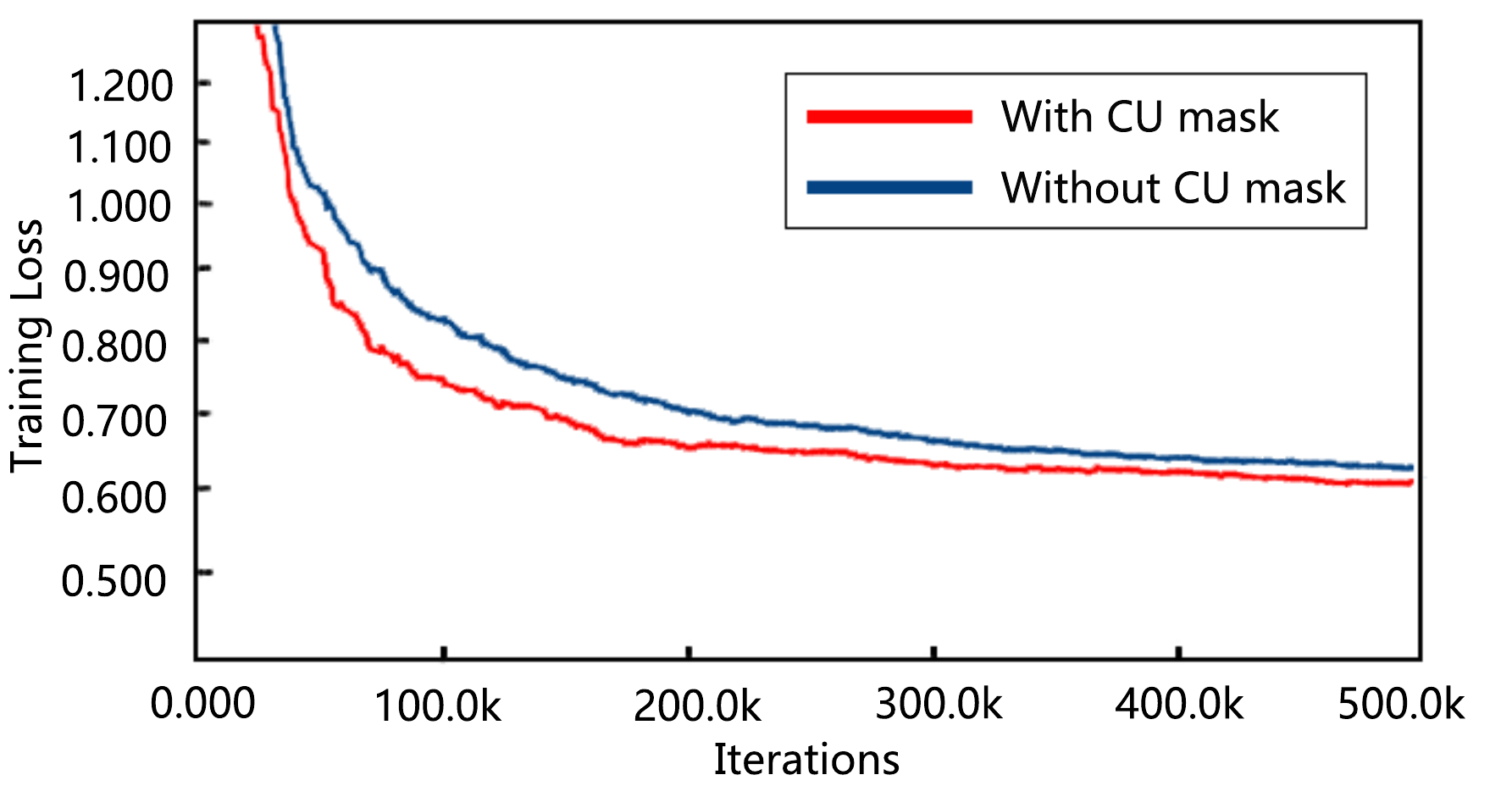}
    \caption{The training loss curves about models with CU mask and without CU mask.}
    \label{fig:w_with_loss_Training}
\end{figure}

\begin{table}[t]
  \centering
  \caption{Encoding and decoding runtime analysis of Ours compared with HM under AI configuration. (seconds/frame)}
    \begin{tabular}{c|c|c|c|c|c|r}
    \hline
    \hline
    \multicolumn{1}{c|}{\multirow{2}[4]{*}{Resolution}} & \multicolumn{3}{c|}{Encoder} & \multicolumn{3}{c}{Decoder} \bigstrut\\
\cline{2-7}          & HM    & Ours & $\triangle T_{enc}$ & HM    & Ours & \multicolumn{1}{c}{$\triangle T_{dec}$} \bigstrut\\
    \hline
    \hline
    4K    & 39.54  & 49.10  & 24\%  & 0.36  & 9.67 & 2620\%\bigstrut\\
    \hline
    6K    & 90.02  & 122.15  & 35\%  & 1.03  & 34.09  & 3412\% \bigstrut\\
    \hline
    8K    & 164.88  & 231.79  & 41\%  & 1.51  & 67.18  & 4432\% \bigstrut\\
    \hline
    \hline
    \end{tabular}%
  \label{tab:times}%
\end{table}%

\begin{table*}[t]
  \centering
  \caption{Ablation studies and analysis in terms of average WS-PSNR (dB) and runtime (second/frame) for different resolution and QPs.}
    \begin{tabular}{c|c|c|c|c|c|c|c|c|c|c|c|c|c|c}
    \hline
    \hline
    \multirow{2}[4]{*}{Method} & \multicolumn{4}{c|}{4K}       & \multicolumn{4}{c|}{6K}       & \multicolumn{4}{c|}{8K}       & \multirow{2}[4]{*}{Average} & Runtime \bigstrut\\
\cline{2-13}\cline{15-15}          & 27 & 32 & 37 & 42   &  27 & 32 & 37 & 42    & 27 & 32 & 37 & 42    &       & 4K \bigstrut\\
    \hline
    Ours(w/o MASK) & 42.14  & 39.62  & 36.92  & 34.15  & 44.06  & 41.18  & 38.26  & 35.28  & 44.84  & 41.96  & 38.91  & 35.78  & 39.42  & 6.64  \bigstrut\\ \hline
    Ours(w/o FS+AR) & 41.97  & 39.56  & 36.89  & 34.08  & 43.83  & 41.10  & 38.20  & 35.20  & 44.62  & 41.88  & 38.83  & 35.70  & 39.32  & 6.96  \bigstrut\\
\cline{2-15}    Ours(w/o FS) & 42.20  & 39.64  & 36.94  & 34.19  & 44.09  & 41.17  & 38.24  & 35.29  & 44.85  & 41.95  & 38.90  & 35.82  & 39.44  & 7.18  \bigstrut\\
\cline{2-15}    Ours(w/o AR) & 42.12  & 39.62  & 36.94  & 34.18  & 44.04  & 41.18  & 38.27  & 35.31  & 44.79  & 41.95  & 38.92  & 35.80  & 39.43  & 6.46  \bigstrut\\
    \hline
    Ours(1 RCAB) & 42.14  & 39.61  & 36.90  & 34.12  & 44.03  & 41.16  & 38.22  & 35.25  & 44.79  & 41.91  & 38.86  & 35.74  & 39.39  & \textbf{4.82}  \bigstrut\\
\cline{2-15}    Ours(2 RCAB) & 42.17  & 39.64  & 36.92  & 34.14  & 44.10  & 41.20  & 38.24  & 35.26  & 44.87  & 41.98  & 38.90  & 35.78  & 39.43  & 5.40  \bigstrut\\
\cline{2-15}    Ours(3 RCAB) & 42.18  & 39.66  & 36.95  & \textbf{34.20}  & 44.10  & 41.21  & 38.26  & 35.31  & 44.86  & 41.98  & 38.93  & 35.83  & 39.46  & 6.08  \bigstrut \\ \hline
    Ours  &\textbf{42.26}  & \textbf{39.70}  & \textbf{36.97}  & \textbf{34.20}  & \textbf{44.26}  & \textbf{41.28}  & \textbf{38.31}  & \textbf{35.32}  & \textbf{45.01}  & \textbf{42.03}  & \textbf{38.95}  & \textbf{35.84}  & \textbf{39.51}   & 6.62  \bigstrut\\
    \hline
    \hline
    \end{tabular}%
  \label{tab:ablation}%
\end{table*}%

\subsection{Ablation Study}\label{sec:ablation}
To validate the contribution of each module in our framework, we conduct ablation studies to demonstrate their influence. Detailed experimental results are provided as follows.

\subsubsection{Feature extraction module}
To verify the mask-guided feature extraction contribution, we replace the mask guided partial convolutional layers with the common 2D convolutional layers (denoted as \textbf{w/o MASK}). As observed from Fig.~\ref{fig:w_with_loss_Training}, the feature extraction module with mask guidance converges faster than that without mask guidance during the training stage. The training loss is also decreased when introducing mask guidance. This is reasonable since the feature maps generated by the mask guided feature extraction contain many zeros, which make the feature sparse and more conducive to the convergence of the network. From the quantitative results, as shown in Table~\ref{tab:ablation}, we can see that the average WS-PSNR value drops from 39.51dB to 39.42dB, indicating that the mask guided feature extraction module is important for performance improvement.

\begin{table}[t]
  \centering
  \caption{The results of Our model with/without $\mathcal{L}_{v}$ compared with HM under configuration of AI for BD-rate (\%) in terms of viewport 0 and viewport 1 for 360$^o$ video sequences (HM coding as anchor).}
    \begin{tabular}{c|c|c|c|c}
    \hline
    \hline
    \multicolumn{1}{c|}{\multirow{2}[4]{*}{AI}} & \multicolumn{2}{c|}{PSNR(VP0)} & \multicolumn{2}{c}{PSNR(VP1)} \bigstrut\\
\cline{2-5}          & w/o $\mathcal{L}_{v}$ & with $\mathcal{L}_{v}$ & w/o $\mathcal{L}_{v}$ & with $\mathcal{L}_{v}$ \bigstrut\\
    \hline
    4K    & -6.6\% & \textbf{-7.4\%} & -6.9\% & \textbf{-7.7\%} \bigstrut\\
    \hline
    6K    & -6.1\% & \textbf{-6.9\%} & -6.3\% & \textbf{-7.0\%} \bigstrut\\
    \hline
    8K    & -6.5\% & \textbf{-7.0\%} & -7.0\% & \textbf{-7.5\%} \bigstrut\\
    \hline
    Average & -6.4\% & \textbf{-7.1\%} & -6.7\% & \textbf{-7.4\%} \bigstrut\\
    \hline
    \hline
    \end{tabular}%
  \label{tab:with_wo_lv}%
\end{table}%

\subsubsection{Feature recalibration module}
To verify the contributions of the layers in the RCAB unit, we evaluate three different variants of the RCAB unit: i) replace the depthwise separative convolution of FS layer by the common 2D convolution (denoted as \textbf{w/o FS}); ii) remove the AR layer (denoted as \textbf{w/o AR}); iii) combine i) and ii) (denoted as \textbf{w/o FS+AR}).

For the setting \textbf{w/o FS}, we can see that the average WS-PSNR value drops from 39.51dB to 39.44dB as shown in Table~\ref{tab:ablation}. It indicates that the depthwise separative convolution can efficiently learn channel-wise feature representation, especially on the performance improvement of frames compressed with the low QP setting. The reason is that for high quality frames, the FS layer can exploit the channel-wise spatial dependency. Besides, the introduction of depthwise separative convolution can also reduce the computational complexity. Given a convolutional layer with  128 channels and $5\times 5$ kernels, the computational complexity ratio of a depthwise separation convolutional layer to a convolutional layer is only $\frac{1}{128}+\frac{1}{5\times 5}=0.048$.

For the setting \textbf{w/o AR}, the average WS-PSNR value drops 0.08dB as shown in Table~\ref{tab:ablation}. With the AR layer, the features are mutually aggregated and discriminative by learning explicit channel interdependencies. For the setting \textbf{w/o FS+AR}, the average WS-PSNR value drops 0.19dB. Since the residual block only has two convolutional layers when the FS layer is replaced and the AR layer is removed simultaneously, it does not readjust the features.

Besides, the influence of the number of RCABs on performance is also evaluated. As shown in Table.~\ref{tab:ablation}, introducing more RCABs will increase the model capacity, resulting in performance improvement with 0.11 dB on average. However, with more RCABs stacked, more parameters are introduced, which increases the computational complexity. During the practical application, the users can adjust the number of RCABs for achieving a trade-off between the computational complexity and R-D performance. 

\subsubsection{Distortion-aware loss}
We perform the ablation study about the model with and without $\mathcal{L}_{v}$ ($\lambda$ = 0). The results are shown in Table \ref{tab:with_wo_lv}, where the model with $\mathcal{L}_{v}$ leading to 0.7\% on average bit rate reduction compares with the model without $\mathcal{L}_{v}$. Therefore, the distortion-aware loss can not only ensure the overall quality of the video, but also improve the quality of the viewpoint images.

\section{Conclusion}
\label{sec:con}
In this paper, we propose a distortion-aware loop filtering model to improve the quality of reconstructed intra equirectangular images. Specifically, we propose a novel feature extraction module to support kernel size adaption with respect to both latitudes and image contents, which can improve model performance and speed up the convergence of the model training. Furthermore, our proposed feature recalibration module can automatically adjust the weights of feature channels to train a single model for various compression-quality images. The proposed distortion-aware optimization combines the WMSE and the perceptual loss to guarantee local FoV and global image qualities of the restored equirectangular image. Experimental results show that our proposed model effectively suppresses compression artifacts.


%





\ifCLASSOPTIONcaptionsoff
  \newpage
\fi



%

\bibliographystyle{IEEEtran}
\bibliography{main}

\begin{thebibliography}{10}
\providecommand{\url}[1]{#1}
\csname url@samestyle\endcsname
\providecommand{\newblock}{\relax}
\providecommand{\bibinfo}[2]{#2}
\providecommand{\BIBentrySTDinterwordspacing}{\spaceskip=0pt\relax}
\providecommand{\BIBentryALTinterwordstretchfactor}{4}
\providecommand{\BIBentryALTinterwordspacing}{\spaceskip=\fontdimen2\font plus
\BIBentryALTinterwordstretchfactor\fontdimen3\font minus
  \fontdimen4\font\relax}
\providecommand{\BIBforeignlanguage}[2]{{%
\expandafter\ifx\csname l@#1\endcsname\relax
\typeout{** WARNING: IEEEtran.bst: No hyphenation pattern has been}%
\typeout{** loaded for the language `#1'. Using the pattern for}%
\typeout{** the default language instead.}%
\else
\language=\csname l@#1\endcsname
\fi
#2}}
\providecommand{\BIBdecl}{\relax}
\BIBdecl

\bibitem{lin19jestcs}
J.~{Lin}, Y.~{Lee}, C.~{Shih}, S.~{Lin}, H.~{Lin}, S.~{Chang}, P.~{Wang},
  L.~{Liu}, and C.~{Ju}, ``Efficient projection and coding tools for 360$^o$
  video,'' \emph{IEEE Journal on Emerging and Selected Topics in Circuits and
  Systems}, vol.~9, no.~1, pp. 84--97, 2019.

\bibitem{Wien19jestcs}
M.~Wien, J.~M. Boyce, T.~Stockhammer, and W.-H. Peng, ``Standardization status
  of immersive video coding,'' \emph{IEEE Journal on Emerging and Selected
  Topics in Circuits and Systems}, vol.~9, no.~1, pp. 5--17, 2019.

\bibitem{sullivan2012overview}
G.~J. Sullivan, J.-R. Ohm, W.-J. Han, and T.~Wiegand, ``Overview of the high
  efficiency video coding ({HEVC}) standard,'' \emph{IEEE Transactions on
  Circuits and Systems for Video Technology}, vol.~22, no.~12, pp. 1649--1668,
  2012.

\bibitem{yang2019low}
H.~Yang, L.~Shen, X.~Dong, Q.~Ding, P.~An, and G.~Jiang, ``Low-complexity ctu
  partition structure decision and fast intra mode decision for versatile video
  coding,'' \emph{IEEE Transactions on Circuits and Systems for Video
  Technology}, vol.~30, no.~6, pp. 1668--1682, 2019.

\bibitem{list2003adaptive}
P.~List, A.~Joch, J.~Lainema, G.~Bjontegaard, and M.~Karczewicz, ``Adaptive
  deblocking filter,'' \emph{IEEE Transactions on Circuits and Systems for
  Video Technology}, vol.~13, no.~7, pp. 614--619, 2003.

\bibitem{norkin2012hevc}
A.~Norkin, G.~Bjontegaard, A.~Fuldseth, M.~Narroschke, M.~Ikeda, K.~Andersson,
  M.~Zhou, and G.~Van~der Auwera, ``{HEVC} deblocking filter,'' \emph{IEEE
  Transactions on Circuits and Systems for Video Technology}, vol.~22, no.~12,
  pp. 1746--1754, 2012.

\bibitem{fu2012sample}
C.-M. Fu, E.~Alshina, A.~Alshin, Y.-W. Huang, C.-Y. Chen, C.-Y. Tsai, C.-W.
  Hsu, S.-M. Lei, J.-H. Park, and W.-J. Han, ``Sample adaptive offset in the
  {HEVC} standard,'' \emph{IEEE Transactions on Circuits and Systems for Video
  Technology}, vol.~22, no.~12, pp. 1755--1764, 2012.

\bibitem{tsai2013adaptive}
C.~Tsai, C.~Chen, T.~Yamakage, I.~S. Chong, Y.~Huang, C.~Fu, T.~Itoh,
  T.~Watanabe, T.~Chujoh, M.~Karczewicz \emph{et~al.}, ``Adaptive loop
  filtering for video coding,'' \emph{IEEE Journal of Selected Topics in Signal
  Processing}, vol.~7, no.~6, pp. 934--945, 2013.

\bibitem{Ma16MM}
S.~Ma, X.~Zhang, J.~Zhang, S.~W. C.~Jia, and W.~Gao, ``Nonlocal in-loop filter:
  The way toward next-generation video coding?'' \emph{IEEE MultiMedia},
  vol.~23, no.~2, pp. 16--26, 2016.

\bibitem{zxf17tcsvt}
X.~{Zhang}, R.~{Xiong}, W.~{Lin}, J.~{Zhang}, S.~{Wang}, S.~{Ma}, and W.~{Gao},
  ``Low-rank-based nonlocal adaptive loop filter for high-efficiency video
  compression,'' \emph{IEEE Transactions on Circuits and Systems for Video
  Technology}, vol.~27, no.~10, pp. 2177--2188, 2017.

\bibitem{zhang2018residual}
Y.~Zhang, T.~Shen, X.~Ji, Y.~Zhang, R.~Xiong, and Q.~Dai, ``Residual highway
  convolutional neural networks for in-loop filtering in {HEVC},'' \emph{IEEE
  Transactions on Image Processing}, vol.~27, no.~8, pp. 3827--3841, 2018.

\bibitem{jia2019content}
C.~Jia, S.~Wang, X.~Zhang, S.~Wang, J.~Liu, S.~Pu, and S.~Ma, ``Content-aware
  convolutional neural network for in-loop filtering in high efficiency video
  coding,'' \emph{IEEE Transactions on Image Processing}, vol.~28, no.~7, pp.
  3343--3356, 2019.

\bibitem{zhang19tcsvt}
S.~{Zhang}, Z.~{Fan}, N.~{Ling}, and M.~{Jiang}, ``Recursive residual
  convolutional neural network-based in-loop filtering for intra frames,''
  \emph{IEEE Transactions on Circuits and Systems for Video Technology}, pp.
  1--1, 2019.

\bibitem{maNNtcsvt}
S.~{Ma}, X.~{Zhang}, C.~{Jia}, Z.~{Zhao}, S.~{Wang}, and S.~{Wanga}, ``Image
  and video compression with neural networks: A review,'' \emph{IEEE
  Transactions on Circuits and Systems for Video Technology}, pp. 1--1, 2019.

\bibitem{pmlr19khasanova19}
R.~Khasanova and P.~Frossard, ``Geometry aware convolutional filters for
  omnidirectional images representation,'' in \emph{Proceedings of the 36th
  International Conference on Machine Learning}, vol.~97, 2019, pp. 3351--3359.

\bibitem{xu2020state}
M.~Xu, C.~Li, S.~Zhang, and P.~Le~Callet, ``State-of-the-art in 360 video/image
  processing: Perception, assessment and compression,'' \emph{IEEE Journal of
  Selected Topics in Signal Processing}, vol.~14, no.~1, pp. 5--26, 2020.

\bibitem{tateno2018distortion}
K.~Tateno, N.~Navab, and F.~Tombari, ``Distortion-aware convolutional filters
  for dense prediction in panoramic images,'' in \emph{European Conference on
  Computer Vision}, 2018, pp. 732--750.

\bibitem{su2019kernel}
Y.-C. Su and K.~Grauman, ``Kernel transformer networks for compact spherical
  convolution,'' in \emph{Proceedings of the IEEE Conference on Computer Vision
  and Pattern Recognition}, 2019, pp. 9442--9451.

\bibitem{dlvc}
D.~Liu, Y.~Li, J.~Lin, H.~Li, and F.~Wu, ``Deep learning-based video coding: A
  review and a case study,'' \emph{ACM Computing Surveys}, vol.~53, no.~1, pp.
  11:1--11:35, 2020.

\bibitem{dai2017convolutional}
Y.~Dai, D.~Liu, and F.~Wu, ``A convolutional neural network approach for
  post-processing in {HEVC} intra coding,'' in \emph{International Conference
  on Multimedia Modeling}, 2017, pp. 28--39.

\bibitem{yang2018enhancing}
R.~Yang, M.~Xu, T.~Liu, Z.~Wang, and Z.~Guan, ``Enhancing quality for {HEVC}
  compressed videos,'' \emph{IEEE Transactions on Circuits and Systems for
  Video Technology}, vol.~29, no.~7, p. 2039–2054, 2019.

\bibitem{yang2018cvpr}
R.~Yang, M.~Xu, Z.~Wang, and T.~Li, ``Multi-frame quality enhancement for
  compressed video,'' in \emph{Proc. CVPR}, 2018, p. 6664–6673.

\bibitem{ma2018residual}
L.~Ma, Y.~Tian, and T.~Huang, ``Residual-based video restoration for hevc intra
  coding,'' in \emph{2018 IEEE Fourth International Conference on Multimedia
  Big Data (BigMM)}, 2018, pp. 1--7.

\bibitem{kang2017multi}
J.~Kang, S.~Kim, and K.~M. Lee, ``Multi-modal/multi-scale convolutional neural
  network based in-loop filter design for next generation video codec,'' in
  \emph{IEEE International Conference on Image Processing (ICIP2017)}, 2017,
  pp. 26--30.

\bibitem{lin2019partition}
W.~Lin, X.~He, X.~Han, D.~Liu, J.~See, J.~Zou, H.~Xiong, and F.~Wu,
  ``Partition-aware adaptive switching neural networks for post-processing in
  hevc,'' \emph{IEEE Transactions on Multimedia}, vol.~22, no.~11, pp.
  2749--2763, 2019.

\bibitem{park2016cnn}
W.-S. Park and M.~Kim, ``{CNN}-based in-loop filtering for coding efficiency
  improvement,'' in \emph{2016 IEEE 12th Image, Video, and Multidimensional
  Signal Processing Workshop (IVMSP)}, 2016, pp. 1--5.

\bibitem{kuanar2018deep}
S.~Kuanar, C.~Conly, and K.~Rao, ``Deep learning based {HEVC} in-loop filtering
  for decoder quality enhancement,'' in \emph{2018 Picture Coding Symposium
  (PCS)}.\hskip 1em plus 0.5em minus 0.4em\relax IEEE, 2018, pp. 164--168.

\bibitem{jia2017spatial}
C.~Jia, S.~Wang, X.~Zhang, S.~Wang, and S.~Ma, ``Spatial-temporal residue
  network based in-loop filter for video coding,'' in \emph{2017 IEEE Visual
  Communications and Image Processing (VCIP)}, 2017, pp. 1--4.

\bibitem{meng2018new}
X.~Meng, C.~Chen, S.~Zhu, and B.~Zeng, ``A new {HEVC} in-loop filter based on
  multi-channel long-short-term dependency residual networks,'' in \emph{2018
  Data Compression Conference}, 2018, pp. 187--196.

\bibitem{ding2019switchable}
D.~Ding, L.~Kong, G.~Chen, Z.~Liu, and Y.~Fang, ``A switchable deep learning
  approach for in-loop filtering in video coding,'' \emph{IEEE Transactions on
  Circuits and Systems for Video Technology}, vol.~30, no.~7, pp. 1871--1887,
  2019.

\bibitem{su2017learning}
Y.-C. Su and K.~Grauman, ``Learning spherical convolution for fast features
  from 360 imagery,'' in \emph{Advances in Neural Information Processing
  Systems}, 2017, pp. 529--539.

\bibitem{zhao2018distortion}
Q.~Zhao, C.~Zhu, F.~Dai, Y.~Ma, G.~Jin, and Y.~Zhang, ``Distortion-aware cnns
  for spherical images,'' in \emph{IJCAI}, 2018, pp. 1198--1204.

\bibitem{qiao2020viewport}
M.~Qiao, M.~Xu, Z.~Wang, and A.~Borji, ``Viewport-dependent saliency prediction
  in 360 video,'' \emph{IEEE Transactions on Multimedia}, 2020.

\bibitem{li2019viewport}
C.~Li, M.~Xu, L.~Jiang, S.~Zhang, and X.~Tao, ``Viewport proposal cnn for 360°
  video quality assessment,'' in \emph{2019 IEEE/CVF Conference on Computer
  Vision and Pattern Recognition}, 2019, pp. 10\,169--10\,178.

\bibitem{partialConv}
G.~Liu, F.~A. Reda, K.~J. Shih, T.-C. Wang, A.~Tao, and B.~Catanzaro, ``Image
  inpainting for irregular holes using partial convolutions,'' in
  \emph{European Conference on Computer Vision}, 2018, pp. 89--105.

\bibitem{ray2018icassp}
B.~Ray, J.~Jung, and M.-C. Larabi, ``A low-complexity video encoder for
  equirectangular projected 360 video content,'' in \emph{Proc. ICASSP}, 2018,
  pp. 1723--1727.

\bibitem{compressedOS}
L.~Zhao, Z.~He, W.~Cao, and D.~Zhao, ``Real-time moving object segmentation and
  classification from hevc compressed surveillance video,'' \emph{IEEE
  Transaction on Circuits and Systems for Video Technology}, vol.~28, no.~6,
  pp. 1346--1357, 2019.

\bibitem{SENetPAMI}
J.~{Hu}, L.~{Shen}, S.~{Albanie}, G.~{Sun}, and E.~{Wu},
  ``Squeeze-and-excitation networks,'' \emph{IEEE Transactions on Pattern
  Analysis and Machine Intelligence}, 2019.

\bibitem{Xception}
F.~Chollet, ``Xception: Deep learning with depthwise separable convolutions,''
  in \emph{CVPR}, 2017, pp. 1251--1258.

\bibitem{WMSE}
Y.~{Sun}, A.~{Lu}, and L.~{Yu}, ``Weighted-to-spherically-uniform quality
  evaluation for omnidirectional video,'' \emph{IEEE Signal Processing
  Letters}, vol.~24, no.~9, pp. 1408--1412, 2017.

\bibitem{snyder1987map}
J.~P. Snyder, \emph{Map projections--A working manual}.\hskip 1em plus 0.5em
  minus 0.4em\relax US Government Printing Office, 1987, vol. 1395.

\bibitem{xiao2012recognizing}
J.~Xiao, K.~A. Ehinger, A.~Oliva, and A.~Torralba, ``Recognizing scene
  viewpoint using panoramic place representation,'' in \emph{2012 IEEE
  Conference on Computer Vision and Pattern Recognition}, 2012, pp. 2695--2702.

\bibitem{CTC}
P.~Hanhart, J.~Boyce, and K.~Choi, ``{JVET} common test conditions and
  evaluation procedures for 360 video,'' in \emph{Joint Video Exploration Team,
  San Diego, CA, USA, Tech. Rep. {JVET-J}1012}, 2018.

\bibitem{boyce2016common}
J.~Boyce, E.~Alshina, A.~Abbas, and Y.~Ye, ``Common test conditions and
  evaluation procedures for 360 video coding,'' in \emph{ISO/IEC JTC1/SC29/WG11
  N16515}, 2016.

\end{thebibliography}

\end{document}